\documentclass[sigconf]{acmart}
\usepackage{booktabs}
\usepackage{bbding}
\usepackage{color}
 
\usepackage{amssymb}
\usepackage{amsmath}
\usepackage{multirow}
\usepackage{xcolor}
\definecolor{DarkGreen}{RGB}{1,100,32} 

\usepackage[normalem]{ulem}
\useunder{\uline}{\ul}{}

\newtheorem{definition}{Definition}
\usepackage{algorithm}
\usepackage{algorithmic}

\AtBeginDocument{%
  }

\setcopyright{acmlicensed}
\copyrightyear{2018}
\acmYear{2018}
\acmDOI{XXXXXXX.XXXXXXX}

\acmConference[Conference acronym 'XX]{Make sure to enter the correct
  conference title from your rights confirmation emai}{June 03--05,
  2018}{Woodstock, NY}
\acmISBN{978-1-4503-XXXX-X/18/06}

\begin{document}

\title{Toward Scalable Graph Unlearning: A Node Influence Maximization based Approach}


\author{Xunkai Li, Bowen Fan, Zhengyu Wu, Zhiyu Li, Rong-Hua Li, Guoren Wang}
\renewcommand{\shortauthors}{Xunkai Li et al.}

\begin{abstract}
    Machine unlearning, as a pivotal technology for enhancing model robustness and data privacy, has garnered significant attention in prevalent web mining applications, especially in thriving graph-based scenarios. 
    However, most existing graph unlearning (GU) approaches face significant challenges due to the intricate interactions among web-scale graph elements during the model training: 
    (1) The gradient-driven node entanglement hinders the complete knowledge removal in response to unlearning requests; 
    (2) The billion-level graph elements in the web scenarios present inevitable scalability issues.
    To break the above limitations, we open up a new perspective by drawing a connection between GU and conventional social influence maximization. 
    To this end, we propose Node Influence Maximization (NIM) through the decoupled influence propagation model and fine-grained influence function in a scalable manner, which is crafted to be a plug-and-play strategy to identify potential nodes affected by unlearning entities. 
    This approach enables offline execution independent of GU, allowing it to be seamlessly integrated into most GU methods to improve their unlearning performance.
    Based on this, we introduce Scalable Graph Unlearning (SGU) as a new fine-tuned framework, which balances the forgetting and reasoning capability of the unlearned model by entity-specific optimizations.
    Extensive experiments on 14 datasets, including large-scale ogbn-papers100M, have demonstrated the effectiveness of our approach.
    Specifically, NIM enhances the forgetting capability of most GU methods, while SGU achieves comprehensive SOTA performance and maintains scalability.
\end{abstract}

\ccsdesc[500]{Computing methodologies~Semi-supervised learning settings}
\ccsdesc[500]{Computing methodologies~Neural networks}

\keywords{Graph Neural Network; Graph Unlearning; Influence Maximization;}

\maketitle

\section{Introduction}
\label{sec: Introduction}
    Recently, with the rapid growth of graph-oriented AI applications, graph neural networks (GNNs) have become a focal point of interest in the web mining community~\cite{kipf2016gcn,Hu2021ahgae,li2024_atp}. 
    During the GNN deployment, the realistic demands for model robustness and data privacy have led to the emergence of a new learning paradigm - graph unlearning (GU). 
    For instance, in graph-based recommendations, eliminating the negative impact of noise caused by malicious transactions and human errors on the model during the training is necessary to enhance robustness~\cite{liu2023gu_application_recommendation1, yuan2023gu_application_recommendation2, xin2024gu_application_recommendation3}.
    The request for privacy is particularly predominant in healthcare networks, and it is essential to comply with patient's requests to be forgotten by removing the corresponding gradient-driven knowledge~\cite{shaik2023gu_application_healthcare1, said2023gu_application_healthcare2, wang2024gu_application_healthcare3}.
    Therefore, the development of GU is imperative.
    
    In general, GU demands real-time responses to delete graph elements (i.e., unlearning entities, UE) per requests during training, and achieves the following objectives:
    (1) \textbf{Model Update}: Modifying original model weights to eliminate the influence of training gradients generated by the UE (\textit{Forgetting Capability});
    (2) \textbf{Inference Protection}: Ensuring that the data removal does not negatively affect predictions for Non-UE (\textit{Reasoning Capability}).
    Compared to unlearning in CV, GU is more challenging due to the complexity of quantifying the influence of UE.
    For high-influenced entities (HIE) that are affected by UE gradients but concealed within the Non-UE, it is crucial to reveal them for complete knowledge removal.


    Recently, numerous GU methods have been proposed from various perspectives, but inherent limitations persist:
    (1) Projection-based methods: Projector~\cite{cong2023projector} and GraphEditor~\cite{cong2022grapheditor} project the original model into a specific subspace via the optimization constraints to eliminate UE influence. 
    \textit{Limitations: They rely on stringent assumptions for linear GNNs, lacking generalizability.}
    (2) Partition-based methods: GraphEraser~\cite{chen2022graph_eraser}, GUIDE~\cite{wang2023guide}, and GraphRevoker~\cite{zhang2024graphrevoker} partition the graph to enable independent learning. 
    This allows for retraining specific partitions in response to unlearning requests without starting from scratch.
    \textit{Limitations: They rely on well-designed partition and output aggregation strategies that hinder deployment.}
    (3) Gradient-based methods: CGU~\cite{chien2022cgu}, CEU~\cite{wu2023ceu_link}, ScaleGUN~\cite{yi2024ScaleGUN} and GST~\cite{pan2023gst_unlearning} follow the $(\epsilon,\delta)$-approximate certified unlearning~\cite{guo2019certifiedML} under linear GNNs. 
    They establish bounds on the gradient residual norm for graph-based scenarios. 
    GIF~\cite{wu2023gif} introduces graph-based influence functions~\cite{koh2017influence_function2} and directly updates the model by the corresponding Hessian matrix.
    \textit{Limitations: They prioritize certified and exact unlearning from the theoretical perspective over practical deployment, leading to inflexible and sub-optimal performance~\cite{mitchell2021mend}.}
    (4) Learning-based methods: GNNDelete~\cite{cheng2023gnndelete}, MEGU~\cite{li2024megu}, and UtU~\cite{tan2024utu_link} directly construct loss functions for UE, HIE, and Non-UE to fine-tune the original model, aiming to obtain a modified model while achieving inference protection. 
    \textit{Limitations: They often lack scalability in their fine-tune mechanisms and limited exploration of the reliable HIE.}
    More discussions can be found in Appendix~\ref{appendix: Systematic Review of GU baselines}

    

\begin{figure*}[t]   
	\centering
    \setlength{\abovecaptionskip}{0.3cm}
    \setlength{\belowcaptionskip}{-0.3cm}
	\includegraphics[width=\linewidth,scale=1.00]{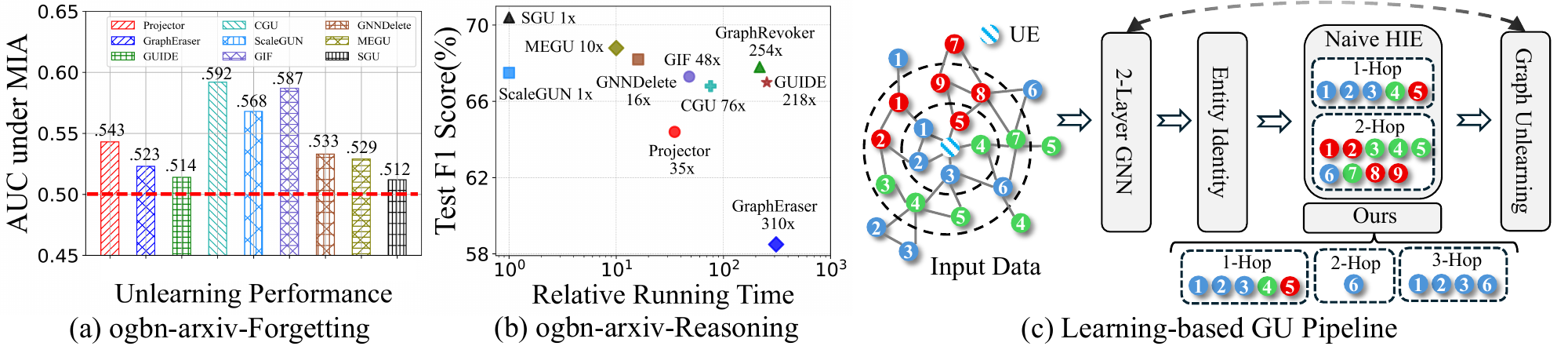}
	\caption{
 (a,b) Performance in the ogbn-arxiv (169k Nodes). 
 The red line (AUC=0.5) represents the best unlearning performance under Membership Inference Attack~\cite{shokri2017mia_attack}.
 (c) Potential limitations of HIE selection under the learning-based GU.}
    \label{fig: motivation}
\end{figure*}

    Reviewing the above GU taxonomies, we conclude that learning-based GU demonstrates \textit{generalizability} across various GNN architectures, potential \textit{scalability} for large-scale scenarios, and a \textit{better trade-off between forgetting and reasoning}. 
    However, the naive HIE selection hinders complete UE knowledge removal. 
    To further illustrate, we present empirical analysis in Fig.~\ref{fig: motivation} along with a toy example highlighting the limitations of existing HIE selection.
    In our experiments, we apply SGC~\cite{wu2019sgc} as the backbone to satisfy the linear assumption of projection- and gradient-based methods and randomly select 10\% training nodes as UE. 
    Notably, we did not choose larger datasets because many GU strategies lack scalability, leading to out-of-memory (OOM) or out-of-time (OOT) errors.

    Based on this, we evaluate SOTA baselines from different GU taxonomies. 
    Specifically, in Fig.\ref{fig: motivation}(a), Membership Inference Attacks (MIA) are utilized to evaluate \textit{Model Update}, quantifying the \textit{forgetting capability}.
    In Fig.\ref{fig: motivation}(b), \textit{Inference Protection} is evaluated by Non-UE prediction, demonstrating \textit{reasoning capability} and running efficiency.
    Our observations are as follows:
    (1) Gradient-based CGU and GIF achieve certified data removal with theoretical guarantees, but their practical deployment is less effective, as claimed by previous studies~\cite{mitchell2021mend,li2024megu}.
    (2) Partition-based GraphEraser, GUIDE, and GraphRevoker exhibit superior forgetting capability due to the retraining of UE-specific partitions. 
    (3) Learning-based MEGU and GNNDelete demonstrate stronger reasoning capability and higher efficiency due to fine-tuned design but leave room for improving forgetting capability. 
    Based on the above observations, we conclude with an assumption that \textit{existing learning-based GU excels in achieving satisfactory predictive ability, but their sub-optimal unlearning performance may result from naive HIE selection strategies.} 
    
    
    To further investigate the learning-based GU pipeline and reveal the inherent limitations of existing HIE selection methods, we present a toy example in Fig.~\ref{fig: motivation}(c).
    Due to the well-known message-passing mechanism, most methods select the $L$-hop neighborhood of UE as the HIE for they are explicitly reached by the UE in the $L$-layer GNN backbone.
    Despite the simplicity and intuitiveness of such a design, the intricate interactions among graph entities during training cannot be adequately captured by the $L$-hop neighborhood.
    Our claims are as follows: 
    (1) Nodes of the same label class as the UE are more likely to be influenced, even beyond the $L$-hop neighborhood, due to the easier transmission of similar gradient patterns during training.
    (2) Nodes of different label classes from the UE are less likely to be influenced unless they are very close to the UE in the topology, as the diluted knowledge over long-distance graph propagation is insignificant compared to gradient norms driven by supervised information. 
    The above claims are inspired by recent advancements in homophily theorems~\cite{zheng2022hete_gnn_survey3, platonov2023hete_gnn_survey4, platonov2022hete_gnn_survey5} and influence-based studies in continual~\cite{choi2024dslr}, imbalance~\cite{chen2021node_reweight}, and active learning~\cite{zhang2021grain}.
    Notably, besides the learning-based GU, other methods such as CGU and GIF also rely on selecting reliable HIE for unlearning. 
    Therefore, developing an appropriate HIE selection strategy that considers the message passing of the GNN is critical. 

    To address the above issues, we introduce a novel perspective that bridges GU with social influence maximization.
    Specifically, we propose Node Influence Maximization (NIM), which treats UE as the seed set and utilizes a decoupled influence propagation model and a fine-grained influence function to identify the node set activated by the seed node set, which is viewed as HIE. 
    This strategy enables a new angle to model the influence of UE during the intricate graph entity interactions in a scalable manner.
    We demonstrate it outperforms traditional neighborhood-based HIE selection (i.e., significantly improves the unlearning performance of HIE-based GU).
    Based on this, to handle unlearning requests for billion-level graphs, we propose Scalable Graph Unlearning (SGU). 
    This approach leverages UE, Non-UE, and reliable HIE obtained by NIM, combined with prototype representation and contrastive learning (CL), to perform lightweight, entity-specific fine-tuning. 
    SGU avoids redundant computations involving most nodes and ensures high running efficiency and scalability. 
    Compared to the most recent scalable GU methods, ScaleGUN, SGU safeguards forgetting and reasoning capability based on efficient fine-tuning (details in Appendix~\ref{appendix: Our Approach and ScaleGUN}).

    \textbf{Our contributions.}
    (1) \textit{\underline{New Perspective.}} 
    To the best of our knowledge, this work is the first to introduce a new perspective for modeling the intricate influence of UE during GU by social influence maximization.
    (2) \textit{\underline{Plug-and-play Strategy.}} 
    We propose NIM, which identifies HIE through a decoupled influence propagation model and a fine-grained influence function in a scalable manner, to seamlessly integrate with any HIE-based GU strategies for improving unlearning performance.
    (3) \textit{\underline{Lightweight Framework.}} 
    We propose SGU, which additionally employs the prototype representation and CL to construct entity-specific optimization objectives, enabling efficient fine-tuning that ensures comprehensive SOTA performance.
    (4) \textit{\underline{SOTA Performance.}}
    Extensive evaluations on 14 datasets prove that NIM has a substantial positive impact on the unlearning performance of prevalent HIE-based GU (up to 5.74\% improvement) and SGU achieves the most complete knowledge removal (up to 7.26\% fuller) and the better inference protection (up to 4.85\% higher).

\section{Preliminaries}
\subsection{Problem Formalization}
\label{sec: Problem Formalization}
    In this work, we primarily utilize semi-supervised node classification to evaluate GU. 
    This task is based on the topology of the labeled set $\mathcal{V}_L$ and the unlabeled set $\mathcal{V}_U$, where the nodes in $\mathcal{V}_U$ are supervised by $\mathcal{V}_L$.
    Furthermore, the core of link-level GU evaluation is to achieve collaborative edge representation using connected nodes.
    Consider an undirected graph $\mathcal{G} = (\mathcal{V}, \mathcal{E}, \mathcal{X})$ with $|\mathcal{V}|=n$ nodes, $|\mathcal{E}|=m$ edges, and $\mathcal{X}=\mathbf{X}$.
    The feature matrix is $\mathbf{X} = \{x_1,\dots,x_n\}$ in which $x_v\in\mathbb{R}^{f}$ represents the feature vector of node $v$, and $f$ represents the dimension of the node attribute information, the adjacency matrix is ${\mathbf{A}}\in\mathbb{R}^{n\times n}$.
    Besides, $\mathbf{Y} = \{y_1,\dots,y_n\}$ represents the label matrix, where $y_v\in\mathbb{R}^{|\mathcal{Y}|}$ denotes a one-hot vector and $|\mathcal{Y}|$ is the number of the label classes.
    In GU, after receiving unlearning request $\Delta \mathcal{G}=\left\{\Delta \mathcal{V}, \Delta \mathcal{E}, \Delta \mathcal{X}\right\}$ on original model parameterized by $\mathbf{W}$, the goal is to modify trainable model weights and output the predictions of Non-UE, both with minimal impact from the UE.
    The typical unlearning requests include feature-level $\Delta \mathcal{G}=\left\{\varnothing,\varnothing,\Delta \mathcal{X}\right\}$, node-level $\Delta \mathcal{G}=\left\{\Delta \mathcal{V},\varnothing,\varnothing\right\}$, edge-level $\Delta \mathcal{G}=\left\{\varnothing,\Delta \mathcal{E},\varnothing\right\}$ in $\mathcal{V}_L$.

\subsection{Scalable Graph Neural Networks}
\label{sec: Scalable Graph Neural Networks}
    Motivated by the spectral graph theory and deep neural networks, GCN~\cite{kipf2016gcn} simplifies the topology-based convolution operator~\cite{2013firstgnn} by the first-order approximation of Chebyshev polynomials~\cite{kabal1986cheby_spectral_graph}.
    The forward propagation of the $l$-th layer GCN is formulated as:
    \begin{equation}
        \label{eq: gcn}
        \mathbf{X}^{(l)} = \sigma(\tilde{\mathbf{A}}\mathbf{X}^{(l-1)}\mathbf{W}^{(l)}),\;\tilde{\mathbf{A}} = \hat{\mathbf{D}}^{-1/2}\hat{\mathbf{A}}\hat{\mathbf{D}}^{-1/2},
    \end{equation}
    where $\hat{\mathbf{D}}$ represents the degree matrix of $\hat{\mathbf{A}}$ (self-loop), $\mathbf{W}$ represents the trainable weights, and $\sigma(\cdot)$ represents the non-linear activation function.
    Intuitively, GCN aggregates the neighbors’ representation embeddings from the $(l\!-\!1)$-th layer to form the representation of the $l$-th layer.
    Such a simple paradigm has proved to be effective in various downstream tasks.
    However, GCN suffers from severe scalability issues since it executes the feature propagation and transformation recursively and is trained in a full-batch manner.
    To achieve scalability, decouple-based approaches have been investigated.

    Recent studies~\cite{zhang2022pasca} have observed that non-linear feature transformation contributes little to performance as compared to graph propagation. 
    For instance SGC~\cite{wu2019sgc} reduces GNNs into a linear model operating on $k$-layer propagated features as follows:
    \begin{equation}
        \label{eq: sgc}
        \begin{aligned}
        \mathbf{X}^{(k)}=\tilde{\mathbf{A}}^k \mathbf{X}^{(0)},\;\mathbf{Y}=\operatorname{softmax}\left(\mathbf{W} \mathbf{X}^{(k)}\right),
        \end{aligned}
    \end{equation}
    where $\mathbf{X}^{(0)}=\mathbf{X}$ and $\mathbf{X}^{(k)}$ is the $k$-layer propagated features.
    As the propagated features $\mathbf{X}^{(k)}$ can be precomputed, SGC is easy to scale to large graphs.
    Inspired by it, SIGN~\cite{frasca2020sign} proposes to concatenate the propagated features in a learnable manner.
    S$^2$GC~\cite{zhu2021ssgc} proposes to average the propagated results from the perspective of spectral analysis $\mathbf{X}^{(k)}=\sum_{l=0}^k \tilde{\mathbf{A}}^l \mathbf{X}^{(0)}$.
    GBP~\cite{chen2020gbp} utilizes the $\beta$ weighted manner to achieve propagation $\mathbf{X}^{(k)}=\sum_{l=0}^k w_l \tilde{\mathbf{A}}^l \mathbf{X}^{(0)}, w_l=\beta(1-\beta)^l$.
    GAMLP~\cite{gamlp} achieves information aggregation based on the attention mechanisms ${\mathbf{X}}^{(k)}=\tilde{\mathbf{A}}^k\mathbf{X}^{(0)} \| \sum_{l=0}^{k-1} w_l \mathbf{X}^{(l)}$, where attention weight $w_l$ has multiple calculation versions.
    GRAND+~\cite{feng2022grand+} proposes a generalized forward push propagation algorithm to obtain $\tilde{\mathbf{P}}$, which is used to approximate $k$-order PageRank weighted $\tilde{\mathbf{A}}$ with higher flexibility.
    Then it obtains propagated results $\tilde{\mathbf{X}}= \tilde{\mathbf{P}}\mathbf{W}\mathbf{X}^{(0)}$.

\subsection{Social Influence Maximization}
\label{sec: Social Influence Maximization}
    The influence maximization (IM) problem in social networks aims to select $\mathcal{B}$ nodes to maximize the number of activated (or influenced) nodes in the network~\cite{kempe2003social_influence_maximization}, which captures the network diffusion dynamics.
    Given a graph $\mathcal{G}=(\mathcal{V}, \mathcal{E})$, the formulation is as follows:

    \begin{equation}
        \label{eq: base im}
        \begin{aligned}
\max _S|\sigma(S)|, \text { s.t. } S \subseteq \mathcal{V},|S|=\mathcal{B},
        \end{aligned}
    \end{equation}
    where $\sigma(S)$ denotes the set of nodes activated by the seed set $S$ under certain influence propagation models, such as the Linear Threshold and Independent Cascade models~\cite{kempe2003social_influence_maximization}.
    The maximization of $\sigma(S)$ is NP-hard. 
    However, if $\sigma(S)$ is non-decreasing and sub-modular with respect to $S$, a greedy algorithm can provide an approximation guarantee of $\left(1-1/e\right)$~\cite{nemhauser1978sim_guarantee}.
    Despite some studies~\cite{yang2012sim_study_2, tang2015sim_study_4, guo2020sim_study_6, ohsaka2020sim_study_7, zhao2021sim_study_8} making significant efforts to optimize algorithm complexity from the perspectives of numerical linear algebra, integrating IM with contemporary graph learning paradigms remains limited. 
    Among the few studies, RIM~\cite{zhang2021rim} and Grain~\cite{zhang2021grain} integrate IM with active learning by using activated nodes to enhance data efficiency. 
    IMBM~\cite{gasteiger2022imbm} utilizes activated nodes and current nodes as a group to empower graph sampling. 
    Scapin~\cite{wang2023scapin} combines IM with adversarial attacks on graphs, aiming to identify and perturb critical graph elements.
    Despite their effectiveness, these methods are designed for specific scenarios and cannot be directly applied to GU. 
    
    In this paper, we focus on bridging reliable HIE selection and NIM within GU rather than addressing classic social IM problems. 
    Specifically, we propose a decoupled-based influence propagation model and a fine-grained influence function for NIM by leveraging the characteristics of decoupled-based scalable GNNs, including both feature and topology influence.
    Based on NIM, we propose SGU, a scalable GU framework that outperforms SOTA baselines in handling various unlearning requests. 
    In a nutshell, NIM and SGU build on the influence propagation in IM, showcasing both the feasibility and potential of this connection and raising a promising future direction.
    More discussions about them are in Appendix~\ref{appendix: Our Proposal and Traditional Social IM}.

\begin{figure*}[t]
	\centering
    \setlength{\abovecaptionskip}{0.1cm}
    \setlength{\belowcaptionskip}{-0.2cm}
  \includegraphics[width=\textwidth]{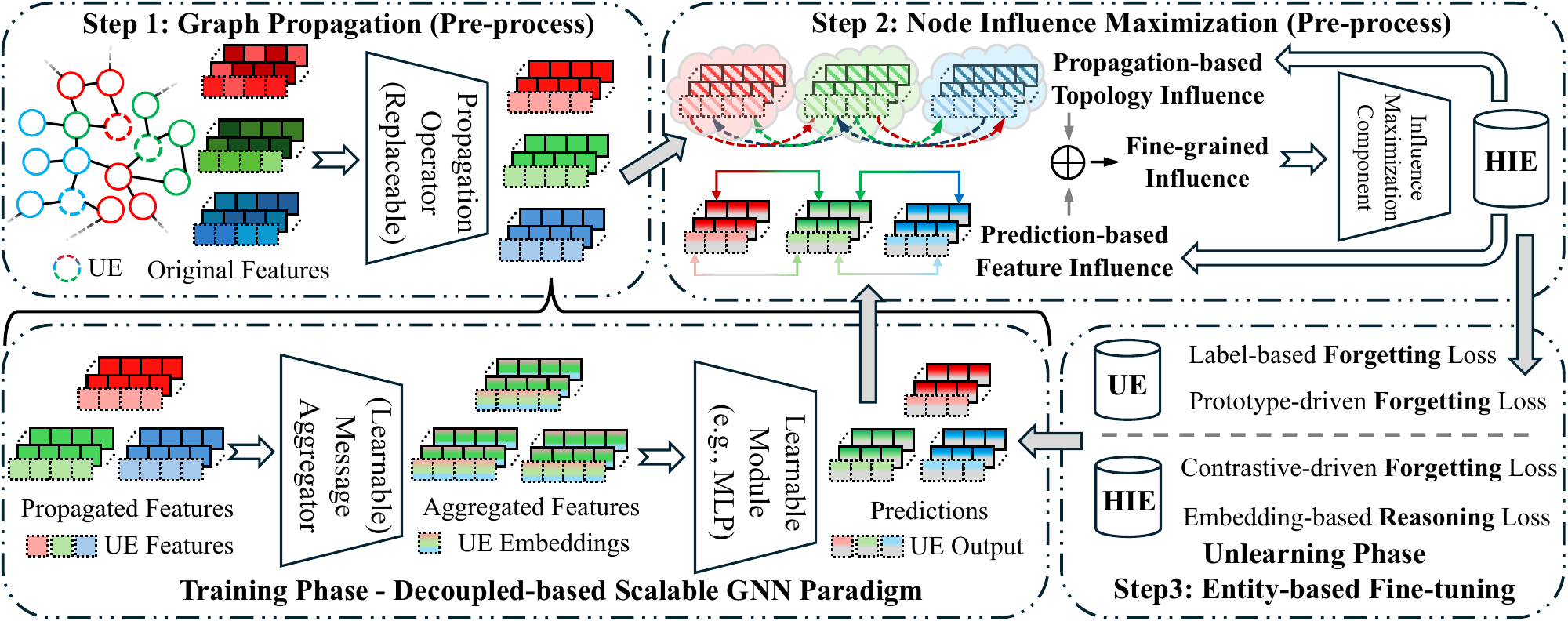}
  \caption{
    Overview of our proposed SGU framework.
}
    \vspace{-0.1cm}
  \label{fig: framework}
\end{figure*}

\section{Scalable Graph Unlearning Framework}
    In this section, we introduce SGU in detail. 
    As depicted in Fig.~\ref{fig: framework}, upon receiving a data removal request, if the current GNN backbone satisfies the decoupled-based paradigm, SGU can be seamlessly integrated with it.
    As for the sampling-based scalable GNNs, SGU requires only an additional weight-free graph propagation.
    In our illustration, SGU first utilizes pre-cached propagated features and prediction to capture the structural and node semantic insights. 
    This enables a fine-grained quantification of the influence of UE on Non-UE from topology and feature perspectives. 
    Subsequently, SGU integrates these influences into a unified criterion and utilizes it to select HIE.
    This process is repeated until the unlearning budget $\mathcal{B}$ is exhausted.
    Based on this, we identify UE and HIE, which are then used to construct entity-specific optimization objectives for computation-friendly fine-tuning.
    This is because it significantly reduces computational dependence on a large number of irrelevant nodes (i.e., Non-UE without HIE) during gradient propagation.
    Meanwhile, this process involves prototype representation and CL, ensuring a favorable trade-off between forgetting and reasoning.
    Now, we introduce each component of SGU in detail.
    The complete algorithm and complexity analysis can be found in Appendix~\ref{appendix: Algorithm and Complexity Analysis}.

\subsection{Node Influence Maximization}
\label{sec: Node Influence Maximization}
    The message-passing mechanism in most GNNs creates intricate interactions among graph entities during training, which results in a complex and unquantifiable gradient flow.
    This poses a significant challenge in completely removing the model knowledge contributed by UE.
    Recent studies have observed that GNNs primarily derive their benefits from weight-free graph propagation rather than learnable module~\cite{zhang2021ndls,zhang2022pasca,li2024_atp}, promoting the decoupled-based scalable GNNs. 
    Notably, this weight-free graph propagation is crucial to the interactions of graph entities and offers a new perspective for analyzing the complex influence of UE without gradient interference.
    Inspired by social influence maximization, we bridge it with GU and establish a new influence quantification criterion.

\subsubsection{Decoupled-based Influence Propagation Model}
\label{sec: Decoupled-based Influence Propagation Model}
    To quantify the intricate interactions, we clarify $k$-step weight-free graph propagation and $\mathbf{W}^\star$-weighted learning process in a decoupled manner:
    \begin{equation}
        \label{eq: weight-free graph propagation}
        \begin{aligned}
        &\tilde{\mathbf{X}}=\mathbf{P}\mathbf{X}=\sum_{l=0}^{k}\sum_{i=0}^l w_i\cdot\left(\hat{\mathbf{D}}^{r-1}\hat{\mathbf{A}}\hat{\mathbf{D}}^{-r}\right)^i\mathbf{X}^{(l)}, \tilde{\mathbf{Y}}=\mathbf{W}^{\star}\tilde{\mathbf{X}},
        \end{aligned}
    \end{equation}
    where $\mathbf{P}$ is the graph propagation equation, serving as the paradigm to model node proximity measures and GNN propagation formulas (i.e., $\tilde{\mathbf{A}}^l\mathbf{X}^{(k-1)}$ in SGC~\cite{wu2019sgc} and $w_l (\hat{\mathbf{D}}^{-a}\hat{\mathbf{A}}\hat{\mathbf{D}}^{-b})^l\mathbf{X}^{(k-1)}$ in AGP~\cite{wang2021agp}). 
    For a given node $u$, a node proximity query yields $\mathbf{P}(v)$, representing the importance of $v$ with respect to $u$. 
    The weight sequence $w_i$ and kernel coefficient $r$ affects transport probabilities for modeling node proximity.
    For more instantiations of $\mathbf{P}$, please refer to Sec.~\ref{sec: Scalable Graph Neural Networks}.

    By taking Eq.~(\ref{eq: weight-free graph propagation}) as a type of influence propagation model, NIM utilizes GNN backbone to obtain $\tilde{\mathbf{X}}$ and $\tilde{\mathbf{Y}}$, achieving model-agnostic topology and feature influence quantification. 
    This is because $\tilde{\mathbf{X}}$ captures structural insights from the $l$-hop neighbors, guided by the probabilities obtained from $k$-step propagation originating from the source node to other nodes in the graph. 
    Meanwhile, $\tilde{\mathbf{Y}}$ establishes the gradient dependencies among nodes through label supervision, capturing potential feature-oriented relationships.
    Now, we define node-wise influence through representation differences as follows:
\begin{definition}{(\textbf{\textit{Topology Influence}}). }
\label{def: Topology Influence}
    The topology influence of node $u$ on node $v$ after $k$-step graph propagation is the L1-norm of the expected Jacobin matrix in propagated features:
    \begin{equation}
    \label{eq: topology influence}
    \begin{aligned}
    I_t(v,u,k)=\left\|\mathbb{E}\left[\partial \tilde{\mathbf{X}}_v^{(k)} / \partial \tilde{\mathbf{X}}_u^{(0)}\right]\right\|_1.
    \end{aligned}
    \end{equation}
\end{definition}
\begin{definition}{(\textbf{\textit{Feature Influence}}). }
\label{def: Feature Influence}
    The feature influence of node $u$ on node $v$ after $k$-step graph propagation is the L1-norm of the expected Jacobin matrix in model predictions:
    \begin{equation}
    \label{eq: feature influence}
    \begin{aligned}
    I_f(v,u,k)=\left\|\mathbb{E}\left[\partial \tilde{\mathbf{Y}}_v^{(k)} / \partial \tilde{\mathbf{Y}}_u^{(0)}\right]\right\|_1.
    \end{aligned}
    \end{equation}
\end{definition}
    Given the $k$-step propagation, $I(v, u, k)$ captures the sum over probabilities of all possible influential paths from $v$ to $u$. 
    Consider propagation operator as $\mathbf{P}=\hat{\mathbf{D}}^{-1}\hat{\mathbf{A}}$, the normalized $\tilde{I}(v, u, k)$ is the probability that random walk starting at $v$ ends at $u$ within $k$ steps:
    \begin{equation}
    \label{eq: influence quantification}
    \begin{aligned}
    \tilde{I}(v, u, k)=\frac{I(v, u, k)}{\sum_{o\in\Delta\mathcal{V}}I(v, o, k)}=\sum_{\mathcal{P}_k^{v \rightarrow u}} \prod_{i=k}^1 \widetilde{a}_{v_{(i-1)}, v(i)},
    \end{aligned}
    \end{equation}
    where $\mathcal{P}_k^{v \rightarrow u}$ be a path $[v_{(k)}, v_{(k-1)}, \ldots, v_{(0)}]$ of length $k$ from node $v$ to $u$ and $\widetilde{a}_{v_{(i-1)}, v_{(i)}}$ is the normalized weight of edge $(v_{(i)}, v_{(i-1)})$.

\subsubsection{Fine-grained Influence Function}
    Intuitively, the weak influence of $u$ on $v$ with small $\tilde{I}(v, u, k)$ would have a limited impact on $v$ due to few influence paths to propagate topology-driven structural insights and feature-driven label supervision.
    Therefore, we define the activated node $v$ by requiring the maximum influence $\tilde{I}(v, S, k)=\max_{u \in S} \tilde{I}(v, u, k)$ of $S$ on the node $v$, which is larger than a threshold.
    In the context of GU, $S$ and its activated node set are the UE and HIE.
    Now, we provide the formal definition below:
\begin{definition}{(\textbf{\textit{IM-based HIE}}). }
\label{def: theta-based HIE}
    Consider node-wise UE derived from the unlearning request as $\Delta \mathcal{V}$, Eq.~(\ref{eq: weight-free graph propagation}) as the influence propagation model.
    The UE is seed node set $S$ and HIE is the activated node set $\sigma(S)$, which is a subset of nodes in $\mathcal{V}/\Delta\mathcal{V}$ and activated by $S$.
    Given a threshold $\theta$, the IM-based HIE (activated node set) is defined as:
    \begin{equation}
    \label{eq: hie}
    \begin{aligned}
    \sigma(S)=\bigcup_{v \in \mathcal{V}/\Delta \mathcal{V}, \tilde{I}(v, S, k)\geq\theta}\{v\}, S = \Delta \mathcal{V}.
    \end{aligned}
    \end{equation}
\end{definition}
    In our implementation, we quantify the fine-grained influence of UE on Non-UE via $\tilde{I}(v,u,k)=\tilde{I}_t(v,u,k) + \tilde{I}_f(v,u,k)$ to identify HIE.
    Compared to Eq.~(\ref{eq: base im}), in the context of GU, the seed set is fixed as UE and the budget $\mathcal{B}$ is the size of HIE. 
    Consequently, we do not need to expend effort in designing IM maximization criteria to discover potential $S$. 
    Instead, we only need to utilize $\tilde{I}(v,u,k)$ to identify $\sigma(S)$ as HIE.
    Notably, larger $\theta$ requires a stronger influence from the UE on Non-UE to activate them as HIE.
    If $\mathcal{B}$ is ample, a smaller $\theta$ is set to identify more HIE to achieve complete forgetting.
    Conversely, with a limited $\mathcal{B}$, a larger $\theta$ should be set to identify HIE within the constrained capacity, prioritizing the efficiency and prediction with less fine-tuning and more original knowledge.
    Therefore, we can flexibly balance efficiency, unlearning, and prediction by $\mathcal{B}$ and $\theta$.

\subsection{Entity-based Fine-tuning}
\label{sec: Entity-based Fine-tuning}
    In general, the key idea of SGU is to implement efficient fine-tuning through entity-specific optimization objectives (i.e., UE and HIE) from the perspectives of forgetting and reasoning. 
    This allows us to edit the original model $\mathbf{W}$ with minimal training overhead ({Model Update}), eliminating the gradient impact of UE while maintaining the predictive performance of Non-UE ( {Inference Protection}). 
    Notably, in large-scale graphs, the size of UE and HIE is significantly smaller than Non-UE, thereby avoiding redundant computations.

\subsubsection{Unlearning Entities Perspective}
\label{sec: Unlearning Entities Perspective}
    To completely remove the gradient-driven model knowledge contributed by the UE, we have the following key insights:
    (1) \textit{Prediction-level}: 
    SGU employs label-driven supervision information to directly erase the model memory of UE predictions, achieving a simple yet intuitive forgetting.
    (2) \textit{Embedding-level}: 
    SGU further removes the model memory of UE embeddings by generating prototypes from other nodes with UE-specific labels, achieving complete forgetting.
    Notably, although Non-UE is used in generating prototype representations, this process does not involve gradient updates and needs to be performed only once before fine-tuning. 
    Therefore, it can be considered a type of self-supervised information with minimal computational cost.

\noindent\textbf{Label-based Forgetting.}
    To achieve simple yet effective forgetting of UE by label-driven supervision, we construct the cross-entropy loss by randomly shuffling $\operatorname{Random}(\cdot)$ the UE-corresponding labels:
\begin{equation}
    \label{eq: ue label-based forgetting1}
    \begin{aligned}
    \mathcal{L}_{f}^1 = -\sum_{u\in\Delta \mathcal{V}}\operatorname{Random}\left(\mathbf{Y}_u^{ij}\right)\log\hat{\mathbf{Y}}_u^{ij},\;\Delta \mathcal{V}=\mathcal{T}\left(\Delta \mathcal{G}\right).
    \end{aligned}
\end{equation}
    To address various types of unlearning requests in Sec.~\ref{sec: Problem Formalization} within NIM, we use a transformation function $\mathcal{T}(\cdot)$ to convert data removal requests into node UE.
    Specifically, for feature and node unlearning, we perform identity mapping. 
    As for edge unlearning, the two nodes connected by the unlearning edges are treated as UE.

\noindent\textbf{Prototype-driven Forgetting.}
    Building upon the above direct prediction-level unlearning, we further achieve comprehensive unlearning by perturbing the UE embedding space through prototype representation. 
    This process can be formally expressed as:
\begin{equation}
    \label{eq: ue prototype-based forgetting2}
    \begin{aligned}
        &\;\;\;\;\;\;\mathcal{P}^c=\frac{1}{\left|\mathcal{S}_{\mathcal{V}/\Delta \mathcal{V}}(c)\right|}\sum_{\left(x_i,y_i\right) \in \mathcal{S}_{\mathcal{V}/\Delta \mathcal{V}}(c)}f_{\operatorname{emb}}\left(x_i\right),\\
        &\mathcal{L}_{f}^2=\sum_{c=1}^{|\mathcal{Y}|}\sum_{(x_u,y_u)\in\Delta \mathcal{V}(c)}\left\Vert{f_{\operatorname{emb}}\left(x_u\right)}-\operatorname{Random}\left(\mathcal{P}^c\right)\right\Vert_F,
    \end{aligned}
\end{equation}
    where $\mathcal{S}$ is the set of samples annotated with label class $c$. 
    Notably, GNN typically comprises embedding and prediction components $f_{\operatorname{emb}}$ and $f_{\operatorname{pre}}$. 
    The former maps initial sample features to an embedding space for semantic extraction, while the latter produces task-specific outputs using these semantic embeddings.
    These prototypes are derived from class-specific averaging embeddings, which encapsulate high-level statistic-conveyed semantic information.
    By constructing the above optimization objective, SGU can separate UE representations from Non-UE representations in the global embedding space, thereby achieving complete knowledge removal.

\subsubsection{High-influence Entities Perspective}
    Generally, removing the gradient knowledge contributed by UE negatively impacts model predictions, particularly for HIE, since the data removal weakens the label-driven supervision and disrupts the model. 
    Enforcing the forgetting of UE will exacerbate this deterioration. 
    This presents a dilemma: 
    \textit{We cannot ensure the complete removal of UE influence for HIE, but meanwhile the decline of HIE prediction seems inevitable.}
    
    To break this limitation, we reformulate the optimization objectives for HIE and have the following key insights:
    (1) \textit{Embedding-level}: 
    SGU additionally integrates Non-UE for sampling, constructing CL loss to completely eliminate UE influence.
    (2) \textit{Prediction-level}: 
    SGU uses memory-based supervision from the original model to ensure predictions.
    Although these two optimization objectives seem to be in conflict, they can coexist harmoniously.
    This is because SGU aims to obtain a new representation perspective for HIE that diverges from the original embedding space without affecting predictions.
    This approach preserves predictive performance while eliminating UE influence by embedding transformation.
    Moreover, although SGU utilizes Non-UE for sampling to provide abundant information, this process is only required once before fine-tuning.

\noindent\textbf{CL-driven Forgetting.}
    Similar to Prototype-driven Forgetting, we aim to perturb the original embedding space of HIE to eliminate UE influence. 
    However, since HIE essentially belongs to Non-UE, we must ensure its predictive performance.
    Inspired by CL, to provide predictive guidance, we perform label-specific sampling on Non-UE (without HIE) as positive query samples for each anchor embedding $h_v$ in HIE. 
    As for UE knowledge removal, we perform label-specific sampling within UE and HIE as negative query samples for $h_v$:
\begin{equation}
\small
\label{eq: ue cl-based forgetting3}
    \begin{aligned}
    \mathcal{L}_{f}^3=-\log\frac{\sum_{u\in\mathcal{S}_{pos}}\exp\left({d\left(h_{v},h_{u}\right)}\right)}{\sum_{u\in\mathcal{S}_{pos}}\exp\left({d\left(h_{v},h_{u}\right)}\right)+\sum_{u\in\mathcal{S}_{neg}}\exp\left(d\left(h_{v},h_{u}\right)\right)},
    \end{aligned}
\end{equation}
    where $d(\cdot)$ is the distance metric between the anchor and query samples.
    Therefore, overall forgetting loss is $\mathcal{L}_f = \mathcal{L}_f^1+\mathcal{L}_f^2+\mathcal{L}_f^3$.

\noindent\textbf{Memory-based Reasoning.}
    Benefiting from the standard training process before receiving data removal requests, the original GNN model weights $\mathbf{W}$ and predictions $\tilde{\mathbf{Y}}$ encapsulate strong reasoning capability. 
    This reliable memory can be viewed as a type of self-supervised information empowering the fine-tuning as follows:
\begin{equation}
    \label{eq: hie memory-based prediction}
    \begin{aligned}
        &\mathcal{L}_{p} \!=\! \operatorname{L2}\left(\mathbf{W}\right) + \operatorname{KL}\left(\tilde{\mathbf{Y}}, \hat{\mathbf{Y}}\right)\!=\!\Vert\mathbf{W}\Vert^2\!+\!\sum_{(i, j)} \tilde{\mathbf{Y}}_{{HIE}}^{i j} \log \frac{\tilde{\mathbf{Y}}_{{HIE}}^{i j}}{\hat{\mathbf{Y}}_{{HIE}}^{i j}} .
    \end{aligned}
\end{equation}
    The first term is L2 regularization, which under the supervision of $\mathcal{L}_f$ aims to limit model updates to preserve reasoning capability under UE knowledge removal.
    The second term is KL loss, which directly leverages the strong reasoning capability of the original model for HIE to supervise the predictions of the fine-tuned model.

\section{Experiments}
    In this section, we conduct extensive experiments on our approach.
    To begin with, we introduce 14 benchmark datasets along with GNN backbones and GU baselines. 
    Subsequently, we present the GU evaluation methodology. 
    Details about these experimental setups can be found in Appendix~\ref{appendix: Dataset Description}-\ref{appendix: Link-specific Evaluation}.
    After that, we aim to address the following questions:
    \textbf{Q1}: As a plug-and-play module, what is the impact of NIM on HIE-based GU?
    \textbf{Q2}: Compared to existing GU strategies, can SGU achieve SOTA performance?
    \textbf{Q3}: If SGU is effective, what contributes to its performance?
    \textbf{Q4}: How robust is SGU when dealing with different scales of unlearning requests and hyperparameters?
    \textbf{Q5}: What is the running efficiency of SGU?

\subsection{Experimental Setup}
\noindent\textbf{Datasets.}
    In the node-specific transductive scenario, we conduct experiments on five citations and three co-purchase networks. 
    In the inductive scenario, we perform experiments on protein, image, and social networks.
    We also consider collaboration, protein, and citation networks to provide a comprehensive evaluation in the link-specific scenario. 
    More details can be found in Appendix~\ref{appendix: Dataset Description}.

\noindent\textbf{Baselines.}
   We conduct experiments using the following backbones:
   (1) Prevalent GNNs: GCN, GAT, GIN;
   (2) Sampling-based GNNs: GraphSAGE, GraphSAINT, ClusterGCN;
   (3) Decouple-based GNNs: SGC, SIGN, GBP, S$^2$GC, AGP, GAMLP;
   (4) Link-specific GNNs: SEAL, NeoGNN, BUDDY, NCNC, MPLP.
   We compare SGU with GraphEditor, GUIDE, GraphRevoker, CGU, CEU, GIF, D2DGN, GNNDelete, UtU, MEGU, ScaleGUN (details in Appendix~\ref{appendix: Compared Baselines}-\ref{appendix: Experiment Environment}).
   OOT occurs when GU requires more runtime than Retrain.

\noindent\textbf{Evaluation.}
    For node-specific scenarios, the evaluation methodologies are as follows:
    (1) \textbf{Model Update}: 
    Inspired by prevalent methods, we set up the following experiments to quantify the unlearning capability:
    (i) \textit{Edge Attack}~\cite{wu2023gif,li2024megu}: 
    We randomly select two nodes with different labels to add noisy edges, treating them as UE. 
    Intuitively, a method that achieves better unlearning will effectively mitigate their negative impact on predictions.
    (ii) \textit{MIA}~\cite{chen2022graph_eraser,wang2023guide,cheng2023gnndelete}: 
    The attacker, with access to both the original and modified model, determines whether a specific node has been revoked from the modified model.
    The evaluation maintains a 1:1 ratio and higher AUC(>0.5) indicates more information leakage.
    (2) \textbf{Inference Protection}:
    We directly evaluate the reasoning capability of the modified model by reporting the Non-UE predictions via the F1 score.
    Please refer to Appendix~\ref{appendix: Link-specific Evaluation} for more details on the link-specific evaluation.

\subsection{Performance Comparison}
\label{sec: Performance Comparison}

\begin{table}[]
\setlength{\abovecaptionskip}{0.2cm}
\setlength{\belowcaptionskip}{-0.2cm}
\caption{NIM improvement on HIE-based GU under MIA.
}
\label{tab: nim improvement}
\resizebox{\linewidth}{25mm}{
\setlength{\tabcolsep}{1.2mm}{
\begin{tabular}{c|cccccc}
\midrule[0.3pt]
GU-Unlearning & Cora  & CiteSeer & PubMed & PPI   & Flickr & Improv.               \\ \midrule[0.3pt]
CGU           & 0.592 & 0.596    & 0.575  & 0.608 & 0.612  & \multirow{2}{*}{\textcolor{black}{$\Uparrow$5.12$\%$}} \\
CGU+NIM       & 0.564 & 0.568    & 0.546  & 0.577 & 0.575  &                       \\ \midrule[0.3pt]
GIF           & 0.587 & 0.604    & 0.580  & 0.613 & 0.614  & \multirow{2}{*}{\textcolor{black}{$\Uparrow$5.28$\%$}} \\
GIF+NIM       & 0.556 & 0.562    & 0.539  & 0.583 & 0.572  &                       \\ \midrule[0.3pt]
D2DGN         & 0.573 & 0.584    & 0.567  & 0.592 & 0.590  & \multirow{2}{*}{\textcolor{black}{$\Uparrow$4.54$\%$}} \\
D2DGN+NIM     & 0.545 & 0.552    & 0.541  & 0.574 & 0.568  &                       \\ \midrule[0.3pt]
GNNDelete     & 0.554 & 0.568    & 0.560  & 0.583 & 0.585  & \multirow{2}{*}{\textcolor{black}{$\Uparrow$4.25$\%$}} \\
GNNDelete+NIM & 0.528 & 0.537    & 0.532  & 0.546 & 0.544  &                       \\ \midrule[0.3pt]
MEGU          & 0.560 & 0.573    & 0.554  & 0.572 & 0.582  & \multirow{2}{*}{\textcolor{black}{$\Uparrow$4.46$\%$}} \\
MEGU+NIM      & 0.532 & 0.540    & 0.527  & 0.537 & 0.551  &                       \\ \midrule[0.3pt]
\end{tabular}
}}
\end{table}

\noindent\textbf{A Hot-and-plug HIE Selection Module.}
    To answer \textbf{Q1}, we present the improvements in forgetting capability brought by integrating NIM into HIE-based GU approaches in Table~\ref{tab: nim improvement}. 
    Notably, we report only the unlearning performance because identifying HIE is essentially about revealing the impact of UE, which aims to complete knowledge removal. 
    This improvement is directly reflected in the unlearning performance. 
    Additionally, we conduct ablation experiments based on SGU in Table~\ref{tab: ablation study} to further validate our claims, where we observe that NIM also brings a slight improvement to the reasoning capability.
    We attribute this improvement to reliable HIE, which prompts the entity-specific optimization objectives and achieves a better trade-off between forgetting and reasoning.

\begin{table}[]
\setlength{\abovecaptionskip}{0.2cm}
\setlength{\belowcaptionskip}{-0.2cm}
\caption{Node-level predictive performance within SGC.
}
\label{tab: sgu node-level feature edge unlearning}
\resizebox{\linewidth}{30mm}{
\setlength{\tabcolsep}{1.2mm}{
\begin{tabular}{c|cccc}
\midrule[0.3pt]
Datasets ($\rightarrow$)  & \multicolumn{2}{c}{Photo} & \multicolumn{2}{c}{Computer} \\ \midrule[0.3pt]
GU Request ($\rightarrow$) & Feature                & Edge                  & Feature                   & Edge         \\ \midrule[0.3pt]
Retrain                   & 89.86±0.4               & 90.13±0.3             & 82.46±0.3                 & 82.14±0.4    \\
GraphEditor               & 85.21±0.2               & 84.84±0.3             & 79.55±0.1                 & 78.72±0.1    \\
GUIDE                     & 77.46±0.1               & 78.65±0.1             & 75.72±0.0                 & 76.24±0.0    \\
GraphRevoker              & 81.25±0.1               & 82.72±0.1             & 77.13±0.0                 & 77.55±0.0    \\
CGU                       & 86.37±0.3               & 85.28±0.4             & 79.10±0.2                 & 78.43±0.3    \\
CEU                       & 84.83±0.4               & \underline{86.95±0.3} & 78.41±0.3                 & 79.86±0.4    \\
GIF                       & 86.59±0.4               & 86.13±0.3             & 80.68±0.3                 & 80.10±0.3    \\
ScaleGUN                  & 86.35±0.5               & 85.94±0.4             & 80.10±0.3                 & 79.47±0.4    \\
GNNDelete                 & 86.90±0.4               & 86.34±0.5             & \underline{80.81±0.3}     & 80.24±0.4    \\
UtU                       & 85.57±0.6               & 86.87±0.6             & 78.14±0.2                 & \underline{80.59±0.3}    \\
MEGU                      & \underline{87.63±0.4}   & 86.80±0.5             & 80.72±0.5                 & 80.45±0.4    \\
SGU                       & \textbf{89.78±0.5}      & \textbf{89.64±0.6}    & \textbf{82.95±0.4}        & \textbf{83.16±0.3}    \\ \midrule[0.3pt]
\end{tabular}
}}
\vspace{-0.2cm}
\end{table}

\begin{table*}[]
\setlength{\abovecaptionskip}{0.2cm}
\setlength{\belowcaptionskip}{-0.2cm}
\caption{Node-level predictive performance under node unlearning. 
The best result is \textbf{bold}.
The second result is \underline{underlined}.
}
\label{tab: sgu nodel-level node unlearning}
\resizebox{\linewidth}{30mm}{
\setlength{\tabcolsep}{1.8mm}{
\begin{tabular}{cc|ccccccccccc}
\midrule[0.3pt]
Backbone ($\downarrow$)  & GU ($\downarrow$)           & Cora              & CiteSeer          & PubMed            & Photo             & Computer          & arxiv             & products          & papers100M & PPI               & Flickr            & Reddit   \\ \midrule[0.3pt]
\multirow{5}{*}{GCN}     & ScaleGUN     & 82.9±0.3          & 73.6±0.1          & 85.1±0.2          & 88.4±0.1          & 83.2±0.1              & 67.8±0.2          & 71.2±0.2               & -        & 54.3±0.1          & 47.9±0.1          & 93.0±0.1      \\
          & GIF                         & 83.2±0.6          & 74.3±0.2          & 85.7±0.1          & \underline{89.1±0.3} & 84.2±0.1          & 68.2±0.4          & OOT               & -        & 54.5±0.2          & 48.2±0.3          & OOT      \\
          & GNNDelete                   & 83.0±0.7          & \underline{75.0±0.3} & 86.0±0.2          & 88.8±0.4          & 83.7±0.2          & 68.7±0.5          & \underline{72.1±0.3} & -        & \underline{55.3±0.3} & 48.6±0.4          & 93.2±0.3 \\
          & MEGU         & \underline{83.8±0.9} & 74.8±0.5          & \underline{86.2±0.1} & 88.7±0.5          & \underline{84.5±0.3} & \underline{69.0±0.5} & 71.8±0.4          & -        & 55.1±0.6          & \underline{49.5±0.5} & \underline{93.6±0.2} \\
          & SGU          & \textbf{85.7±0.8} & \textbf{76.3±0.4} & \textbf{87.5±0.3} & \textbf{90.4±0.3} & \textbf{86.6±0.3} & \textbf{71.3±0.6} & \textbf{73.5±0.5} & -   & \textbf{58.2±0.5} & \textbf{51.4±0.7} & \textbf{94.1±0.3} \\ \midrule[0.3pt]
\multirow{5}{*}{GraphSAGE} & ScaleGUN   & 82.4±0.2          & 73.8±0.1          & 85.3±0.1          & 88.1±0.2          & 84.2±0.1          & 68.1±0.1          & 72.1±0.2               & 62.4±0.2         & 54.7±0.2          & 48.6±0.2          & 92.9±0.1       \\
          & GIF                         & 82.8±0.5          & 73.7±0.3          & \underline{86.1±0.1} & 88.5±0.5          & \underline{84.6±0.2} & 68.6±0.3          & OOT               & OOM        & 54.8±0.3          & 48.4±0.2          & OOT      \\
          & GNNDelete    & 82.4±0.8          & 74.5±0.4          & 85.8±0.2          & \underline{88.9±0.5} & 84.0±0.3          & \underline{68.9±0.4} & 72.0±0.4          & OOM        & 55.0±0.4          & 47.9±0.5          & 93.0±0.2 \\
          & MEGU         & \underline{83.1±0.6} & \underline{75.2±0.5} & 85.5±0.2          & 88.8±0.8          & 84.4±0.4          & 68.5±0.7          & \underline{72.5±0.5} & OOM        & \underline{55.5±0.5} & \underline{49.2±0.6} & \underline{93.7±0.3} \\
          & SGU          & \textbf{85.4±0.7} & \textbf{76.1±0.6} & \textbf{87.9±0.4} & \textbf{90.6±0.5} & \textbf{87.0±0.5} & \textbf{71.5±0.5} & \textbf{73.9±0.6} & \textbf{64.5±0.5}   & \textbf{57.9±0.4} & \textbf{51.8±0.5} & \textbf{94.4±0.2} \\ \midrule[0.3pt]
\multirow{5}{*}{GAMLP}     & ScaleGUN   & 83.8±0.6          & 75.1±0.2          & 86.7±0.2          & 89.2±0.2          & 85.3±0.1          & 69.6±0.2          & 72.8±0.2               & 62.8±0.3         & 55.7±0.1          & 49.5±0.1          & 93.2±0.0      \\
          & GIF                         & \underline{84.1±0.7} & 75.6±0.3          & 86.9±0.1          & 89.8±0.4          & 85.0±0.3          & 69.8±0.5          & OOT               & OOM        & 55.3±0.3          & 49.3±0.2          & OOT      \\
          & GNNDelete    & 83.5±1.1          & \underline{76.2±0.6} & 87.0±0.3          & 89.5±0.6          & 84.9±0.2          & 69.5±0.4          & OOM               & OOM        & 55.6±0.5          & 49.6±0.4          & OOM      \\
          & MEGU         & 83.9±1.2          & 76.0±0.8          & \underline{87.5±0.2} & \underline{90.2±0.5} & \underline{85.4±0.5} & \underline{70.4±0.8} & \underline{73.4±0.4} & OOM        & \underline{55.9±0.7} & \underline{50.1±0.4} & \underline{94.0±0.3} \\
          & SGU          & \textbf{86.9±0.9} & \textbf{77.5±0.7} & \textbf{89.1±0.2} & \textbf{91.3±0.6} & \textbf{87.9±0.4} & \textbf{72.8±0.7} & \textbf{75.0±0.5} & \textbf{65.2±0.5}   & \textbf{58.8±0.6} & \textbf{52.7±0.6} & \textbf{94.8±0.3} \\ \midrule[0.3pt]
\end{tabular}
}}
\end{table*}

\begin{table}[]
\setlength{\abovecaptionskip}{0.2cm}
\setlength{\belowcaptionskip}{-0.cm}
\caption{Edge-level performance within edge unlearning.
}
\label{tab: sgu edge-level edge unlearning}
\resizebox{\linewidth}{24mm}{
\setlength{\tabcolsep}{3mm}{
\begin{tabular}{cc|ccc}
\midrule[0.3pt]
\multirow{2}{*}{GU ($\downarrow$)} & \multirow{2}{*}{Backbone($\downarrow$)} & collab    & ppa       & citation2 \\
                                   &                                         & HR@50     & HR@100    & MRR       \\ \midrule[0.3pt]
\multirow{5}{*}{UtU}               & SEAL                                    & 63.53±0.34 & 49.80±0.52 & 85.44±0.54 \\
                                   & NeoGNN                                  & 55.28±0.41 & 49.56±0.67 & 86.18±0.95 \\
                                   & BUDDY                                   & 65.13±0.72 & 50.24±0.56 & 87.25±0.42 \\
                                   & NCNC                                    & 65.76±0.66 & 59.73±0.84 & 88.03±0.67 \\
                                   & MPLP                                    & 66.40±0.59 & 61.75±0.78 & 89.62±0.35 \\ \midrule[0.3pt]
\multirow{5}{*}{SGU}               & SEAL                                    & 65.72±0.54 & 51.33±0.43 & 88.05±0.41 \\
                                   & NeoGNN                                  & 58.95±0.61 & 52.27±0.65 & 87.84±0.79 \\
                                   & BUDDY                                   & 66.35±0.73 & 51.59±0.42 & 88.46±0.68 \\
                                   & NCNC                                    & 67.20±0.55 & 62.28±0.56 & 89.47±0.55 \\
                                   & MPLP                                    & 67.59±0.67 & 63.31±0.71 & 90.12±0.48 \\ \midrule[0.3pt]
\end{tabular}
}}
\vspace{-0.4cm}
\end{table}

\noindent\textbf{A New Scalable GU Framework.}
    To answer \textbf{Q2} from the reasoning perspective, we report the node-level predictive performance in Table~\ref{tab: sgu node-level feature edge unlearning} and Table~\ref{tab: sgu nodel-level node unlearning}, which validates that SGU consistently outperforms baselines across all datasets on unlearning requests. 
    For instance, on the Photo and Computer, SGU exhibits a remarkable average improvement of 2.7\% over the SOTA baselines.
    Notably, the persistent advantage in both transductive and inductive settings demonstrates the strong generalization of SGU.
    Meanwhile, Table~\ref{tab: sgu edge-level edge unlearning} provides results on link-level reasoning capacity, which demonstrate that SGU outperforms the most competitive UtU tailored for edge unlearning, achieving average performance gains of 2.1\% across three datasets. 
    These observations indicate that SGU can achieve satisfactory performance on our diversified selections of backbones, highlighting its potential for widespread application.
    
\begin{figure*}[t]   
	\centering
    \setlength{\abovecaptionskip}{0.1cm}
    \setlength{\belowcaptionskip}{-0.3cm}
	\includegraphics[width=\linewidth,scale=1.00]{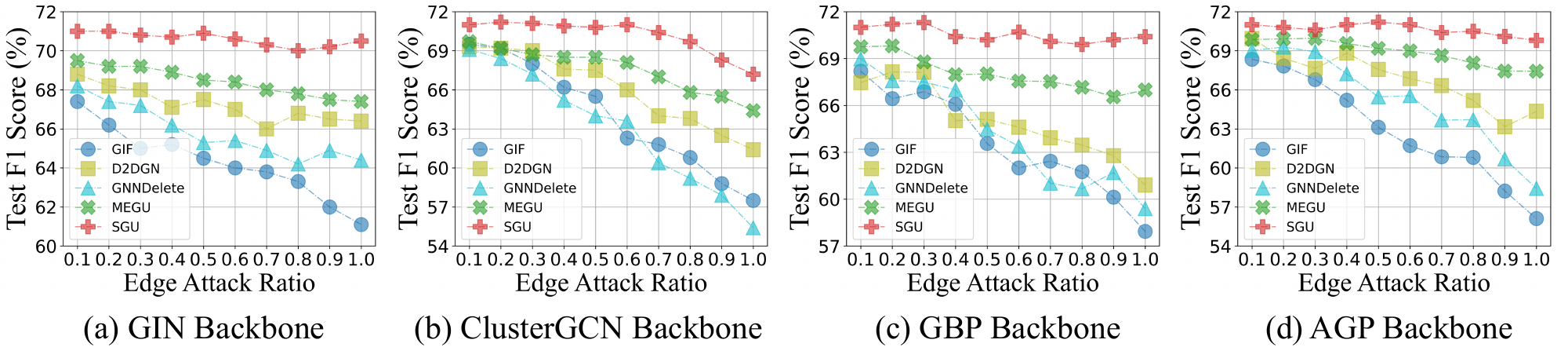}
	\caption{
    Node-level performance on arxiv within Edge Attack.
    The x-axis is the ratio of noisy edges to the existing edges.
    }
    \label{fig: edge attack}
\end{figure*}

    To answer \textbf{Q2} from the forgetting perspective, we visualize the node-level unlearning performance of SGU and competitive baselines (confirmed by previous experiments) in Fig.~\ref{fig: edge attack}. 
    Intuitively, if a GU approach possesses robust forgetting capability, it can mitigate the adverse effects caused by noisy edges, thereby ensuring consistent and satisfactory predictive performance. 
    From the experimental results, we observe that GNNDelete and GIF do not consistently achieve optimal performance. 
    Although D2DGN and MEGU exhibit relative stability, they are still weaker than SGU.
    Due to space limitations, please refer to Appendix~\ref{appendix: Link-specific Evaluation} for more experiments about link-specific prediction and unlearning evaluation.

\subsection{Ablation Study}

\begin{table}[]
\setlength{\abovecaptionskip}{0.2cm}
\setlength{\belowcaptionskip}{-0.cm}
\caption{Ablation study under MIA.
}
\label{tab: ablation study}
\resizebox{\linewidth}{24mm}{
\setlength{\tabcolsep}{2.8mm}{
\begin{tabular}{cc|cccc}
\midrule[0.3pt]
\multirow{2}{*}{Backbone ($\downarrow$)} & \multirow{2}{*}{Module ($\downarrow$)} & \multicolumn{2}{c}{PubMed} & \multicolumn{2}{c}{Computer} \\
                                         &                                        & Unlearn.   & Pred.  & Unlearn.   & Pred.   \\ \midrule[0.3pt]
\multirow{4}{*}{GAT}                     & w/o NIM                                & 0.546        & 87.2±0.6    & 0.550         & 86.8±0.6     \\
                                         & w/o Proto.                             & 0.537        & 87.5±0.3    & 0.546         & 86.8±0.4     \\
                                         & w/o CL Loss                            & 0.529        & 86.8±0.5    & 0.538         & 86.0±0.4     \\
                                         & SGU                                    & 0.518        & 87.6±0.4    & 0.532         & 87.0±0.5     \\ \midrule[0.3pt]
\multirow{4}{*}{SAINT}                   & w/o NIM                                & 0.543        & 87.6±0.3    & 0.566         & 86.1±0.5     \\
                                         & w/o Proto.                             & 0.538        & 87.5±0.3    & 0.554         & 86.4±0.3     \\
                                         & w/o CL Loss                            & 0.535        & 86.3±0.4    & 0.547         & 85.4±0.5     \\
                                         & SGU                                    & 0.524        & 87.8±0.2    & 0.541         & 86.5±0.4     \\ \midrule[0.3pt]
\multirow{4}{*}{SIGN}                    & w/o NIM                                & 0.539        & 87.7±0.3    & 0.557         & 87.2±0.4     \\
                                         & w/o Proto.                             & 0.526        & 87.8±0.2    & 0.540         & 87.0±0.3     \\
                                         & w/o CL Loss                            & 0.517        & 87.0±0.4    & 0.532         & 86.3±0.5     \\
                                         & SGU                                    & 0.510        & 88.1±0.3    & 0.525         & 87.4±0.3     \\ \midrule[0.3pt]
\end{tabular}
}}
\vspace{-0.3cm}
\end{table}

    To answer \textbf{Q3}, we investigate the contributions of NIM, prototype-driven forgetting (Proto.), and CL-driven forgetting (CL Loss) in SGU.
    For NIM, it quantifies the impact of UE on Non-UE during training through a decoupled-based influence propagation model and fine-grained influence function, identifying HIE for constructing subsequent optimization objectives (technical details in Sec.~\ref{sec: Node Influence Maximization}). 
    It can be considered as a pre-processing step for SGU.
    This strikes a better trade-off between forgetting and reasoning, thereby improving the comprehensive performance of GU. 
    For more analysis on NIM, please refer to Sec.~\ref{sec: Performance Comparison}.
    For Proto., which encapsulates high-level, statistic-conveyed semantic information. 
    Based on this, SGU performs prototype perturbation to separate UE from Non-UE representations in the global embedding space, achieving complete knowledge removal (technical details in Eq.(\ref{eq: ue prototype-based forgetting2})). 
    Experimental results in Table~\ref{tab: ablation study} highlight Proto.'s effectiveness in enhancing forgetting capability. 
    Notably, since Eq.(\ref{eq: ue prototype-based forgetting2}) is used solely to enhance unlearning performance, its impact on predictive performance is minimal.
    For CL Loss, it achieves comprehensive improvement by pulling Non-UE and class-specific positive query samples closer while pushing entity-specific negative query samples further in the embedding space, leveraging self-supervised information to guide prediction and transform the embedding space. 
    Results in Table~\ref{tab: ablation study} validate all the above claims on CL. 
    For instance, in the PubMed dataset using GraphSAINT as the backbone, the F1 Score increases from 86.3\% to 87.8\%, and the AUC decreases from 0.535 to 0.524.

\begin{figure*}[t]
    \centering
        \setlength{\abovecaptionskip}{0.15cm}
    \setlength{\belowcaptionskip}{-0.1cm}
    \includegraphics[width=\textwidth]{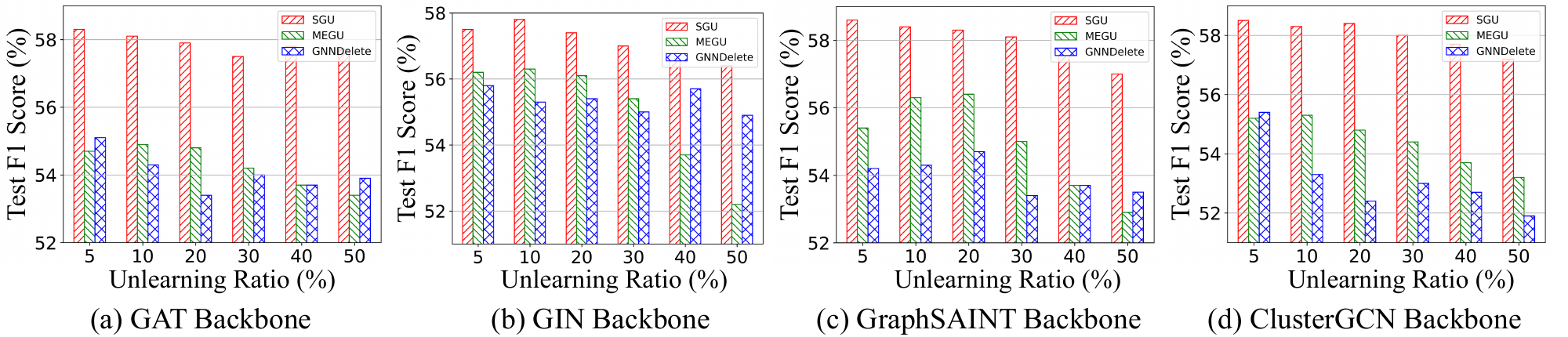}
    \caption{Predictive performance of node unlearning within different ratios on PPI.}
    \label{fig: ratio_node_ppi}
\end{figure*}

\begin{figure}[t]   
	\centering
    \setlength{\abovecaptionskip}{0.1cm}
    \setlength{\belowcaptionskip}{-0.2cm}
	\includegraphics[width=\linewidth,scale=1.00]{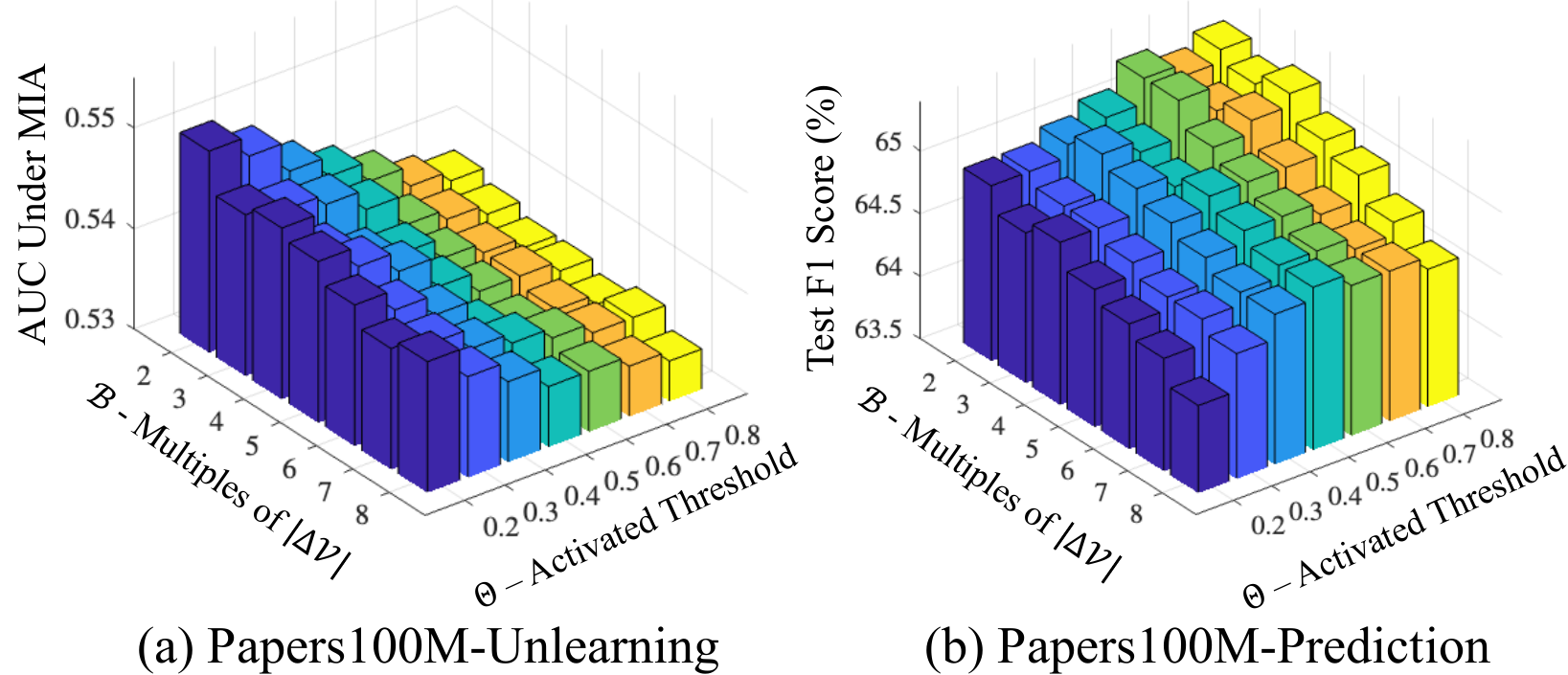}
	\caption{
    Sensitive analysis within GAMLP and node removal.
    }
    \label{fig: exp_hyperparameter}
\end{figure}

\begin{figure}[t]   
	\centering
    \setlength{\abovecaptionskip}{0.1cm}
    \setlength{\belowcaptionskip}{-0.2cm}
	\includegraphics[width=\linewidth,scale=1.00]{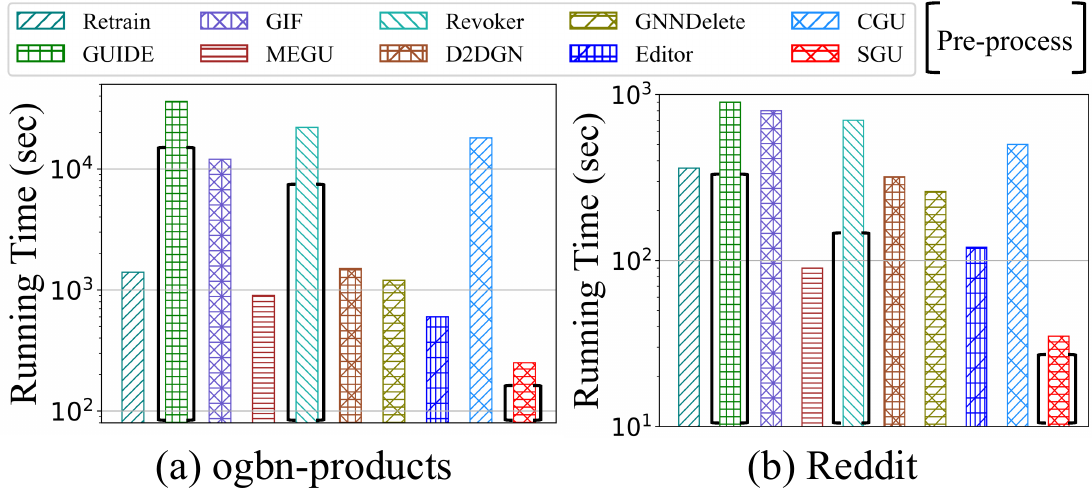}
	\caption{
    Running efficiency within S$^2$GC and node removal.
    }
    \label{fig: exp_efficient}
\end{figure}

\subsection{Robustness Analysis}
\label{sec: Robustness Analysis}

    To answer \textbf{Q4}, we first evaluate SGU and the two most competitive baselines across varying unlearning scales in Fig.~\ref{fig: ratio_node_ppi}. 
    Due to space limitations, additional experimental results and analysis are presented in Appendix~\ref{appendix: Unlearning Challenges at Different Scales}. 
    Our experiments indicate that SGU excels in handling unlearning requests with different scales. 
    Notably, as the unlearning ratio increases, there is an inevitable decline in the performance of the unlearned model across all GU methods. 
    This decline is due to the higher proportion of data removal, which significantly impacts the model predictions.
    However, it is evident that SGU outperforms GNNDelete and MEGU under identical unlearning conditions. 
    These empirical findings and analyses underscore SGU’s robustness in effectively handling unlearning requests.
    
    Subsequently, we investigate $\mathcal{B}$ and $\theta$, and present their sensitivity analysis results for SGU in Fig.~\ref{fig: exp_hyperparameter} from the forgetting and reasoning perspectives.
    In our implementation, NIM utilizes $\mathcal{B}$ and $\theta$ to control the size of HIE and the criteria for selecting HIE. 
    Since HIE is directly related to the entity-specific optimization objectives, its size and quantity determine the extent of fine-tuning and the trade-off between forgetting and reasoning loss (technical details in Sec.~\ref{sec: Entity-based Fine-tuning}).
    Therefore, $\mathcal{B}$ and $\theta$ essentially determine SGU's running efficiency while balancing unlearning and prediction. 
    Our experimental results indicate that both optimal forgetting and reasoning capabilities stem from larger $\theta$ and are associated with larger and smaller $\mathcal{B}$.
    This is because larger $\theta$ strictly filters Non-UE, ensuring high-quality HIE for the subsequent optimization.
    Regarding $\mathcal{B}$, larger values achieve complete knowledge removal and optimal unlearning performance, while smaller values highlight the model's inherent reasoning capability.
    For more details on balancing unlearning and predictive performance by combining the forgetting and reasoning loss through $\lambda$, please refer to Appendix~\ref{appendix: flexible Entity-specific Optimization}.

\subsection{Efficiency Comparison}

    To answer \textbf{Q5}, we present a visualization in Fig.~\ref{fig: exp_efficient} to illustrate the running efficiency. 
    In our report, the running time encompasses both one-time pre-processing and model updating.
    Notably, since the S$^2$GC is essentially a scalable GNN, the computational cost of training from scratch is significantly reduced, setting a higher efficiency benchmark for Retrain.
    Based on this, we draw the following conclusions:
    (1) Partition-based GUIDE and GraphRevoker suffer from excessive time costs during large-scale graph partitioning.
    The computation cost of gradient-based CGU and GIF in large-scale scenarios is unacceptable due to considering all data samples contributing to the gradient for optimization. 
    (2) Projection-based GraphEditor demonstrates high efficiency, attributed to its efficient subspace projection strategy. 
    However, the strict linear assumption hampers its generalizability in deployment. 
    Furthermore, this advantage is not observed on Reddit.
    (3) The most computation cost of SGU is pre-processing, including NIM and sampling for Eq.~(\ref{eq: ue prototype-based forgetting2}) and Eq.~(\ref{eq: ue cl-based forgetting3}). 
    Excluding pre-processing (NIM), the fine-tuning requires minimal time due to entity-specific design. 
    Additionally, since pre-processing is a one-time step, SGU is suitable for real-world scenarios with frequent unlearning requests, offering practical potential.

\section{Conclusion}
    In this paper, we advocate a novel perspective for GU by connecting it with social IM.
    To this end, we propose NIM, which defines a decoupled-based influence propagation model to seamlessly integrate with graph propagation formulas and utilizes a fine-grained influence function as a unified criterion. 
    Based on this, SGU achieves efficient model fine-tuning by entity-specific optimization.
    To further improve NIM, exploring the GU-tailored influence propagation model and quantification functions is worthwhile. 
    Meanwhile, exploring new techniques such as knowledge distillation to design forgetting and reasoning loss is also a promising direction.
\begin{acks}
To Robert, for the bagels and explaining CMYK and color spaces.
\end{acks}

\newpage
\balance{
\bibliographystyle{ACM-Reference-Format}
\bibliography{sample-base}
}
\clearpage
\appendix

\section{Outline}
\begin{description}
    \item[A.1] Systematic Review of GU baselines.
    \item[A.2] Our Approach and ScaleGUN.
    \item[A.3] Our Approach and Traditional Social IM.
    \item[A.4] Algorithm and Complexity Analysis.
    \item[A.5] Dataset Description.
    \item[A.6] Compared Baselines.
    \item[A.7] Hyperparameter settings.
    \item[A.8] Experiment Environment.
    \item[A.9] Link-specific Evaluation.
    \item[A.10] Unlearning Challenges at Different Scales.
    \item[A.11] $\lambda$-flexible Entity-specific Optimization.
\end{description}

\begin{table*}[t]
\caption{A systematic summary of recent GU studies. 
}
\footnotesize
\label{tab: gu_methods}
\begin{tabular}{ccccccc}
\midrule[0.3pt]
Methods                & Type       & Model Agnostic & Request Agnostic & Efficient \& Direct Model Update & Inference Protection & Billion-level Scalability \\ \midrule[0.3pt]
Projector~\cite{cong2023projector} (AISTATS'23) & Projection & \textcolor{red}{\XSolidBrush}            & \textcolor{DarkGreen}{\Checkmark}              & \textcolor{DarkGreen}{\Checkmark}                              & \textcolor{red}{\XSolidBrush}                  & \textcolor{red}{\XSolidBrush}                       \\
GraphEditor~\cite{cong2022grapheditor} (arXiv'23) & Projection & \textcolor{red}{\XSolidBrush}            & \textcolor{DarkGreen}{\Checkmark}              & \textcolor{DarkGreen}{\Checkmark}                              & \textcolor{red}{\XSolidBrush}                  & \textcolor{red}{\XSolidBrush}                       \\
GraphEraser~\cite{chen2022graph_eraser} (CCS'22)   & Partition  & \textcolor{DarkGreen}{\Checkmark}            & \textcolor{DarkGreen}{\Checkmark}              & \textcolor{red}{\XSolidBrush}                              & \textcolor{DarkGreen}{\Checkmark}                  & \textcolor{red}{\XSolidBrush}                       \\
GUIDE~\cite{wang2023guide} (USENIX'23)      & Partition  & \textcolor{DarkGreen}{\Checkmark}            & \textcolor{DarkGreen}{\Checkmark}              & \textcolor{red}{\XSolidBrush}                              & \textcolor{DarkGreen}{\Checkmark}                  & \textcolor{red}{\XSolidBrush}                       \\
GraphRevoker~\cite{zhang2024graphrevoker} (WWW'24)  & Partition  & \textcolor{DarkGreen}{\Checkmark}            & \textcolor{DarkGreen}{\Checkmark}              & \textcolor{red}{\XSolidBrush}                              & \textcolor{DarkGreen}{\Checkmark}                  & \textcolor{red}{\XSolidBrush}                       \\
CGU~\cite{chien2022cgu} (ICLR'23)          & Gradient   & \textcolor{red}{\XSolidBrush}            & \textcolor{DarkGreen}{\Checkmark}              & \textcolor{DarkGreen}{\Checkmark}                              & \textcolor{red}{\XSolidBrush}                  & \textcolor{red}{\XSolidBrush}                       \\
CEU~\cite{wu2023ceu_link} (KDD'23)           & Gradient   & \textcolor{red}{\XSolidBrush}            & \textcolor{red}{\XSolidBrush}              & \textcolor{DarkGreen}{\Checkmark}                              & \textcolor{red}{\XSolidBrush}                  & \textcolor{red}{\XSolidBrush}                       \\
GST~\cite{pan2023gst_unlearning} (WWW'23)           & Gradient   & \textcolor{red}{\XSolidBrush}            & \textcolor{red}{\XSolidBrush}              & \textcolor{DarkGreen}{\Checkmark}                              & \textcolor{red}{\XSolidBrush}                  & \textcolor{red}{\XSolidBrush}                       \\
GIF~\cite{wu2023gif} (WWW'23)           & Gradient   & \textcolor{DarkGreen}{\Checkmark}            & \textcolor{DarkGreen}{\Checkmark}              & \textcolor{DarkGreen}{\Checkmark}                              & \textcolor{red}{\XSolidBrush}                  & \textcolor{red}{\XSolidBrush} \\
ScaleGUN~\cite{yi2024ScaleGUN} (arXiv'24)           & Gradient   & \textcolor{DarkGreen}{\Checkmark}            & \textcolor{DarkGreen}{\Checkmark}              & \textcolor{DarkGreen}{\Checkmark}                              & \textcolor{red}{\XSolidBrush}                  & \textcolor{DarkGreen}{\Checkmark}  \\ 
D2DGN~\cite{sinha2023d2dgn} (arXiv'24)       & Learning   & \textcolor{DarkGreen}{\Checkmark}            & \textcolor{DarkGreen}{\Checkmark}              & \textcolor{DarkGreen}{\Checkmark}                              & \textcolor{DarkGreen}{\Checkmark}                  & \textcolor{red}{\XSolidBrush}                       \\
GCU~\cite{chien2022cgu} (ICBD'23)          & Learning   & \textcolor{DarkGreen}{\Checkmark}            & \textcolor{red}{\XSolidBrush}              & \textcolor{red}{\XSolidBrush}                              & \textcolor{DarkGreen}{\Checkmark}                  & \textcolor{red}{\XSolidBrush}                       \\
GNNDelete~\cite{cheng2023gnndelete} (ICLR'23)    & Learning   & \textcolor{DarkGreen}{\Checkmark}            & \textcolor{DarkGreen}{\Checkmark}              & \textcolor{red}{\XSolidBrush}                              & \textcolor{DarkGreen}{\Checkmark}                  & \textcolor{red}{\XSolidBrush}                       \\
UtU~\cite{tan2024utu_link} (WWW'24)           & Learning   & \textcolor{DarkGreen}{\Checkmark}            & \textcolor{red}{\XSolidBrush}              & \textcolor{red}{\XSolidBrush}                              & \textcolor{DarkGreen}{\Checkmark}                  & \textcolor{red}{\XSolidBrush}                       \\
MEGU~\cite{li2024megu} (AAAI'24)         & Learning   & \textcolor{DarkGreen}{\Checkmark}            & \textcolor{DarkGreen}{\Checkmark}              & \textcolor{DarkGreen}{\Checkmark}                              & \textcolor{DarkGreen}{\Checkmark}                  & \textcolor{red}{\XSolidBrush}                       \\ 
SGU (This Paper)       & Learning   & \textcolor{DarkGreen}{\Checkmark}            & \textcolor{DarkGreen}{\Checkmark}              & \textcolor{DarkGreen}{\Checkmark}                              & \textcolor{DarkGreen}{\Checkmark}                  & \textcolor{DarkGreen}{\Checkmark}                       \\ \midrule[0.3pt]
\end{tabular}
\end{table*}

\subsection{Systematic Review of GU baselines}
\label{appendix: Systematic Review of GU baselines}
    GU remains a burgeoning field with numerous research gaps. 
    To advance its future development and highlight the motivation of our approach, we conduct a systematic review of most existing GU strategies according to our proposed novel taxonomies in Table~\ref{tab: gu_methods}.

    \noindent\textbf{Model Agnostic.}
    It is well-known that quantifying the impact of UE on Non-UE is considerably more challenging in graph-based scenarios than in computer vision due to the intricate interactions among graph entities during model training.
    This complexity makes UE-corresponding knowledge removal difficult, especially in the complex GNN architectures where gradient flows are well hidden.
    To ensure certified or exact data removal with a strict theoretical guarantee, some existing GU strategies simplify the default model architecture to a linear-based GNN to satisfy necessary assumptions, designing the corresponding unlearning algorithms through exhaustive derivations.
    Although they achieve satisfactory theoretical results, the inherent generalization limitations of linear models persist, preventing effective deployment in practical applications.

    \noindent\textbf{Request Agnostic.}
    The inherent complexity of graph entities leads to various data removal requests in GU, as described in Sec.~\ref{sec: Problem Formalization}. 
    To achieve better unlearning and predictive performance in specific graph-based data removal scenarios, some methods design GU algorithms tailored to particular graph entity deletion requests (e.g. edge unlearning). 
    Despite their effectiveness, developing a general-purpose GU remains imperative. 
    Additionally, investigating the potential connections among graph entities to reveal more effective GU paradigms is pivotal to improving comprehensive performance.

    \noindent\textbf{Efficient \& Direct Model Updates.}
    As the most important target of GU, {Model Update} aims to modify the original model parameterized by $\mathbf{W}$ into a new model parameterized by $\mathbf{W}^\star$. 
    This modification ensures that $\mathbf{W}^\star$ forgets the gradient-driven knowledge contributed by UE during model training and eliminates the influence of UE on Non-UE. 
    To achieve this goal, existing GU strategies have proposed various model update mechanisms. 
    We propose that the following methods, which possess Efficient \& Direct characteristics, are more advantageous for real-world deployment:
    (1) Projection-based methods \textit{directly} project the original model weights into specific sub-spaces through optimized constraints.
    (2) Gradient-based methods \textit{directly} update trainable parameters based on gradient convergence analysis.
    (3) Learning-based methods (fine-tuning) \textit{directly} modify the original model using fine-tuning mechanisms.
    These methods \textit{efficiently} achieve {Model Update}, allowing the original model to continue training, offering greater flexibility. 
    
    In contrast, the following methods have limitations:
    (1) Partition-based methods require retraining specific partitions and a partition output aggregation module.
    \textit{Limitations: These methods make them difficult to deploy effectively in practical applications when deletion requests are frequent.}
    (2) Learning-based methods (deletion module) propose freezing the original model and injecting an additional trainable module to achieve unlearning. 
    \textit{Limitations: These methods do not update the original model, preventing further training.}

    \noindent\textbf{Inference Protection.}
    Although the aforementioned projection-based and gradient-based methods achieve Efficient \& Direct {Model Updates} by directly modifying the original model weights, they inadvertently neglect Non-UE during the execution of GU, leading to sub-optimal unlearning and predictive performance. 
    As a result, despite significant theoretical advancements and certifiable, their practical performance is concerning, as highlighted in a recent study~\cite{mitchell2021mend}. 
    In contrast, while partition-based and learning-based methods still have room for improvement in terms of forgetting, these methods place significant emphasis on the predictive performance of Non-UE, thereby achieving {Inference Protection}.
    In a nutshell, considering the deployment of GU in complex real-world applications, their practical unlearning and predictive performance are pivotal in designing efficient GU algorithms.

    \noindent\textbf{Billion-level Scalability.}
    Considering the strict optimization constraints of projection-based methods, the fine-grained graph partitioning and learnable partition output aggregation module of partition-based methods, and the gradient-based methods' reliance on bounds on the gradient residual norm and the computation of the inverse Hessian matrix, these approaches inevitably lack scalability when dealing with billion-level industrial graphs. 
    In contrast, learning-based methods, through carefully designed fine-tuning mechanisms, can be easily extended to large-scale graphs.
    Meanwhile, it ensures the performance of Model Update and Inference Protection while enjoying high running efficiency.

    Based on this, we suggest that GU method should be capable of being applicable to any GNN backbone model (Model Agnostic) and handling any unlearning request at any time (Request Agnostic).
    Hence, it should not only have the ability to adjust the trained model and continue training efficiently and directly (Efficient \& Direct Model Update) but also generate predictions that prioritize the performance of Non-UE (Inference Protection). 
    Furthermore, considering real-world deployment requirements, such methods should demonstrate high efficiency in both training and inference, particularly in industry scenarios (Billion-level Scalability). 
    We hope that the above design principles will drive the future development of GU and inspire more practical approaches.

\subsection{Our Approach and ScaleGUN}
\label{appendix: Our Approach and ScaleGUN}
    Scalable GU has recently gained significant attention.
    ScaleGUN~\cite{yi2024ScaleGUN} represents an important advancement, offering a certifiable mechanism to efficiently handle billion-scale graphs through lazy local propagation, ensuring certified removal of nodes, edges, and features. 
    However, our approach—NIM combined with SGU—introduces critical improvements, particularly in inference protection.
    While ScaleGUN focuses on updating the embedding matrix and emphasizes scalability and certification, SGU addresses both scalability and the challenge of gradient-driven node entanglement. 
    This entanglement complicates complete knowledge removal due to persistent interactions among graph elements.
    NIM decouples influence propagation, identifying and minimizing the impact of unlearned nodes more effectively.
    The key distinction between SGU and ScaleGUN is in inference protection. 
    While ScaleGUN ensures certified unlearning, it does not safeguard post-unlearning model integrity. 
    SGU, through fine-tuning, preserves accurate and reliable predictions, crucial in applications where model integrity is as important as unlearning itself. 
    Additionally, NIM is a plug-and-play strategy with more flexibility, seamlessly integrating with existing GU methods for enhanced performance.
    In summary, SGU offers a more comprehensive solution, addressing inference protection, and ensuring unlearning requests do not degrade model accuracy or reasoning, making it a robust choice for real-world applications.

\subsection{Our Approach and Traditional Social IM}
\label{appendix: Our Proposal and Traditional Social IM}
    The core of our approach in this paper is the NIM module, inspired by the traditional social IM problem. 
    It aims to identify Non-UE significantly influenced by UE during model training as reliable HIE. 
    This guides SGU in constructing entity-specific optimization objectives to achieve complete knowledge removal while preserving predictions. Notably, despite the numerous connections between NIM and social IM, significant differences also exist. 
    Therefore, in this section, we review the traditional social IM problem and the key design of NIM to avoid confusion and further elucidate the intuition behind our approach and potential future improvements.

    As described in Sec.~\ref{sec: Social Influence Maximization}, the traditional social IM problem views the seed node set $S$ as the source of influence, where other nodes that are significantly influenced become $\theta$-based threshold activated node set $\sigma(S)$.
    Specifically, traditional social IM aims to find the seed set $S$ that maximizes $\sigma(S)$ in the current network topology under the following constraints: 
    (1) Certain influence propagation model, such as the Linear Threshold and Independent Cascade models; 
    (2) Quantification function based on threshold $\theta$ to obtain a unified criterion;
    (3) Budget $\mathcal{B}$ decides the size of the seed node set  $|S|=\mathcal{B}$. 
    Although the problem of finding $S$ to maximize $\sigma(S)$ under these constraints is NP-hard, some studies~\cite{nemhauser1978sim_guarantee} have indicated that if $\sigma(S)$ is non-decreasing and sub-modular with respect to $S$, a greedy algorithm can provide an approximation guarantee of $\left(1-1/e\right)$. 
    Consequently, current studies in traditional social IM seek to design more reasonable quantification functions and reduce the theoretical complexity of greedy search processes through numerical linear algebra and algorithmic approaches.
    In a nutshell, in traditional social IM, obtaining the optimal seed set $S$ is the ultimate goal of the optimization problem, while maximizing the activated node set $\sigma(S)$ is one of the constraints. 
    Current studies focus on exploring the potential relationships between quantification criteria and optimization constraints to tighten error bounds through numerical computation methods and improve time and space complexity.
    
    Building upon these foundations, we describe social IM in the context of GU. 
    Although concepts such as the seed node set $S$, activated node set $\sigma(S)$, and influence propagation functions exist in NIM, they have different meanings compared to traditional social IM. 
    Specifically, NIM uses UE to represent the seed node set $S=\Delta\mathcal{V}$, HIE to represent the activated node set $\sigma(S)$, and GNN propagation formulas to represent the influence propagation model (technical details in Sec.~\ref{sec: Decoupled-based Influence Propagation Model}).
    Based on this, the significant differences between NIM and traditional social IM are as follows:
    (1) NIM has a fixed seed node set, thus, the budget $\mathcal{B}$ in GU determines $\sigma(S)$, independent of $S$.
    (2) Unlike the optimization problem addressed in traditional social IM, NIM directly identifies nodes meeting the criteria as $\sigma(S)$ through straightforward calculations based on the unlearning budget $\mathcal{B}$ and threshold $\theta$ using the seed set $S$, the influence propagation model, and the influence quantitative function.
    Notably, since NIM does not involve complex algorithmic problems and challenging optimization constraints, it focuses more on designing suitable influence propagation models and constructing fine-grained influence quantification criteria, ensuring a high-quality and reliable activated node set $\sigma(S)$ as HIE. 
    
    In our implementation, to seamlessly integrate NIM with any reasonable GNN backbone and reduce application complexity, we default to using the propagation formulas in the GNN backbone as the influence propagation model and directly measure the differences in smooth features and soft labels before and after propagation and learnable neural architecture (e.g., MLP) as the influence quantification criterion.
    While experimental results demonstrate that this strategy is simple and effective, a promising direction involves designing a unified influence propagation model specifically suited for GU by considering the common characteristics of graph propagation equations. 
    Additionally, exploring more diverse quantification criteria, such as incorporating neighborhood information when calculating influence, is worth further investigation.
    In a nutshell, although NIM is inspired by traditional social IM, there are significant differences in their definitions and research focuses. 
    However, essentially, they share many commonalities. 
    Extracting key insights from traditional social IM to enhance NIM is valuable.

\begin{algorithm}[t]
\caption{Scalable Graph Unlearning} 
\label{alg: sgu}
\begin{algorithmic}[1] 
\STATE Model training related to downstream tasks;
\STATE Receive graph element deletion request $\Delta\mathcal{G}$;
\STATE Initialize seed set (HIE) $\mathcal{S}=\emptyset$;
\STATE Execute unlearning entity transformation based on Eq.~(\ref{eq: ue label-based forgetting1});
\IF{$f_{prop}$ not satisfies decoupled-based GNN paradigm}
\FOR{$i=1,2,\dots,k$}
\STATE Execute graph propagation to obtain $\tilde{\mathbf{X}}^{k}$ based on Eq.~(\ref{eq: weight-free graph propagation}); 
\ENDFOR
\ENDIF
\STATE Execute inference to obtain soft label $\tilde{\mathbf{Y}}^{k}$ based on the $\tilde{\mathbf{X}}^{k}$;
\FOR{$t=1,2,\dots,\mathcal{B}$}
\FOR{$v\in\mathcal{V}/\Delta\mathcal{V}$}
\STATE Execute the influence quantification function based on the $\tilde{\mathbf{X}}^{k}$, $\tilde{\mathbf{Y}}^{k}$, and Eq.~(\ref{eq: topology influence}-\ref{eq: influence quantification});
\STATE Update $\sigma(\mathcal{S}\cup\{v\})$ for each Non-UE based on the $\tilde{I}(v,u,k)=\tilde{I}_t(v,u,k) + \tilde{I}_f(v,u,k)$; 
\ENDFOR
\STATE $v^\star=\arg\max_{v\in\mathcal{V}/\Delta\mathcal{V}} \tilde{I}(v,u,k)$;
\STATE $\mathcal{S}=\mathcal{S}\cup\{v^\star\}$;
\STATE Obtain UE, HIE, and Non-UE based on the $\mathcal{V}$, $\Delta\mathcal{V}$, and $\mathcal{S}$;
\STATE Execute entity-based fine-tuning based on Eq.~(\ref{eq: ue label-based forgetting1}-\ref{eq: hie memory-based prediction}) to obtain the modified backbone model $f_{gnn}^\star$;
\STATE Execute inference to obtain Non-UE prediction $\tilde{\mathbf{Y}}$;
\RETURN $f_{gnn}^\star$ and $\tilde{\mathbf{Y}}$;
\ENDFOR
\end{algorithmic}
\end{algorithm}
\subsection{Algorithm and Complexity Analysis}
\label{appendix: Algorithm and Complexity Analysis}
    For a more comprehensive overview, we present the complete SGU algorithm in Algorithm~\ref{alg: sgu}. 
    To illustrate its complexity, we use SGC~\cite{wu2019sgc} as the backbone for $k$-step graph propagation. 
    Notably, our analysis can be easily extended to any backbone GNN architecture.
    For a $k$-layer SGC with batch size $b$, the propagated feature $\mathbf{X}^{(k)}$ has a time and space complexity of $O(kmf)$ and $O((b+k)f)$. 
    For a more detailed analysis of the theoretical algorithmic complexity of propagation mechanisms shown in Sec.~\ref{sec: Scalable Graph Neural Networks}, we recommend referring to related studies~\cite{li2024lightdic} that provide comprehensive insights on this topic.
    At this stage, we have captured influence propagation features from a topological perspective and generated soft label predictions from a feature-based perspective using forward inference, with negligible computational overhead.
    Next, we focus on selecting activated nodes as HIEs based on influence measurement, involving matrix computations for distance metrics. 
    We optimize this process using locality-sensitive hashing, an approximate nearest neighbors algorithm, alongside parallelized CPU and GPU computations implemented with NumPy. 
    The time and space complexity of this step are bounded by $O(n \log f / p)$ and $O(n f)$, respectively.
    Finally, we perform fine-tuning by accurately identifying UE, HIE, and Non-UE entities. 
    Due to entity-specific optimizations implemented for this step, the training overhead is minimal compared to the earlier NIM process and can be disregarded.

\subsection{Dataset Description}
\label{appendix: Dataset Description}

\begin{table*}[htbp]
\caption{The statistical information of the experimental datasets.
}
\label{tab: datasets}
\begin{tabular}{ccccccc}
\midrule[0.3pt]
Dataset         & \#Nodes     & \#Features & \#Edges       & \#Classes & \#Task Type     & Description       \\ \midrule[0.3pt]
Cora            & 2,708       & 1,433      & 5,429         & 7         & Transductive    & Citation Network  \\
CiteSeer        & 3,327       & 3,703      & 4,732         & 6         & Transductive    & Citation Network  \\
PubMed          & 19,717      & 500        & 44,338        & 3         & Transductive    & Citation Network  \\ \midrule[0.3pt]
Amazon Photo    & 7,487       & 745        & 119,043       & 8         & Transductive    & Co-purchase Graph \\
Amazon Computer & 13,381      & 767        & 245,778       & 10        & Transductive    & Co-purchase Graph \\ \midrule[0.3pt]
ogbn-arxiv      & 169,343     & 128        & 2,315,598     & 40        & Transductive    & Citation Network  \\
ogbn-products   & 2,449,029   & 100        & 61,859,140    & 47        & Transductive    & Co-purchase Graph \\
ogbn-papers100M & 111,059,956 & 128        & 1,615,685,872 & 172       & Transductive    & Citation Network  \\ \midrule[0.3pt]
PPI             & 56,944      & 50         & 818,716       & 121       & Inductive       & Protein Network   \\
Flickr          & 89,250      & 500        & 899,756       & 7         & Inductive       & Image Network     \\
Reddit          & 232,965     & 602        & 11,606,919    & 41        & Inductive       & Social Network    \\ \midrule[0.3pt]
ogbn-collab     & 235,868	  & 128        & 1,285,465	   & -         & Link Prediction & Collaboration network   \\ 
ogbn-ppa        & 576,289     & 58         & 30,326,273    & -         & Link Prediction & Protein Network   \\ 
ogbn-citation2  & 2,927,963	  & 128        & 30,561,187	   & -         & Link Prediction & Citation Network   \\ \midrule[0.3pt]
\end{tabular}
\vspace{0.1cm}
\end{table*}

    The description of all datasets is listed below:

    \noindent\textbf{Cora}, \textbf{CiteSeer}, and \textbf{PubMed}~\cite{Yang16cora} are three citation network datasets, where nodes and edges represent papers and citation relationships. 
    The node features are word vectors, where each element indicates the presence or absence of each word in the paper.

    \noindent\textbf{Amazon Photo} and \textbf{Amazon Computers}~\cite{shchur2018amazon_datasets} are segments of the Amazon co-purchase graph, where nodes represent items and edges represent that two goods are frequently bought together. 
    Given product reviews as bag-of-words node features.

    \noindent\textbf{ogbn-arxiv} and \textbf{ogbn-papers100M}~\cite{hu2020ogb} are two citation graphs indexed by MAG~\cite{wang2020microsoft_MAG}.
    Each paper involves averaging the embeddings of words in its title and abstract.
    The embeddings of individual words are computed by running the skip-gram model.

    \noindent\textbf{ogbn-products}~\cite{hu2020ogb} is a co-purchasing network, where nodes represent products and edges represent that two products are frequently bought together. 
    Node features are generated by extracting bag-of-words features from the product descriptions.

    \noindent\textbf{PPI}~\cite{zeng2019graphsaint} stands for protein-protein interaction network, where nodes represent protein.  
    If two proteins participate in a life process or perform a certain function together, it is regarded as an interaction between these two proteins. 
    Complex interactions between multiple proteins can be described by PPI networks.
    
    \noindent\textbf{Flickr}~\cite{zeng2019graphsaint} originates from SNAP.
    In this graph, each node represents an image. 
    An edge exists between two nodes if the corresponding images share common properties, such as the same geographic location or the same gallery. 
    Node features are represented by a 500-dimensional bag-of-words model of the images. 
    The labels consist of 81 tags for each image, manually merged into 7 classes, with each image belonging to one of these classes.

    \noindent\textbf{Reddit}~\cite{hamilton2017graphsage} dataset collected from Reddit, where 50 large communities have been sampled to build a post-to-post graph, connecting posts if the same user comments on both.
    For features, off-the-shelf 300-dimensional GloVe vectors are used.

     \noindent\textbf{ogbn-collab}~\cite{wang2020microsoft_MAG} is an undirected graph, representing a subset of the collaboration network between authors. 
     Each node represents an author and edges indicate the collaboration between authors.
     All edges are associated with the year (meta-information), representing the number of co-authored papers published in that year.
     
     \noindent\textbf{ogbn-ppa}~\cite{szklarczyk2019ogbppa} is an undirected, unweighted graph where nodes represent proteins from 58 different species, and edges indicate biologically meaningful associations between proteins.
     Each node is associated with a 58-dimensional one-hot feature vector that indicates the species of the corresponding protein.

     \noindent\textbf{ogbl-citation2}~\cite{wang2020microsoft_MAG} represents the citation network. 
     Each node is a paper with 128-dimensional word2vec features that summarize its title and abstract, and each directed edge indicates that one paper cites another.
     All nodes also come with meta-information indicating the year the corresponding paper was published.

\subsection{Compared Baselines}
\label{appendix: Compared Baselines}
    To evaluate the effectiveness of various GU strategies, we have selected commonly used GNNs as the backbone models to simulate scenarios where unlearning requests are received during training. 
    These compared models represent successful recent designs in scalable graph learning widely applicable in both transductive and inductive settings and link-specific GNNs for link-level downstream tasks. 
    Furthermore, various backbone GNNs can be employed to assess the generalization capability of diverse GU approaches.
    The salient characteristics of all baseline models are outlined below:
    
    \noindent\textbf{GCN}~\cite{kipf2016gcn} introduces a novel approach to graphs that is based on a first-order approximation of spectral convolutions on graphs.
    This approach learns hidden layer representations that encode both local graph structure and features of nodes.

    \noindent\textbf{GAT}~\cite{velivckovic2017gat} utilizes attention mechanisms to quantify the importance of neighbors for message aggregation.
    This strategy enables implicitly specifying different weights to different nodes in a neighborhood, without depending on the graph structure upfront.

    \noindent\textbf{GIN}~\cite{xu2018gin} presents a theoretical framework analyzing GNNs' expressive power to capture graph structures. 
    They develop a simple architecture that is probably the most expressive among GNNs and as powerful as the Weisfeiler-Lehman graph isomorphism test. 

    \noindent\textbf{GraphSAGE}~\cite{hamilton2017graphsage} leverages neighbor node attribute information to efficiently generate representations.
    This method introduces a general inductive framework that leverages node feature information to generate node embeddings for previously unseen data.

    \noindent\textbf{GraphSAINT}~\cite{zeng2019graphsaint} is an inductive framework that enhances training efficiency through graph sampling.
    In each iteration, a complete GCN is built from the properly sampled subgraph, which decouples the sampling from the forward and backward propagation.

    \noindent\textbf{Cluster-GCN}~\cite{chiang2019cluster-gcn} is designed for training with stochastic gradient descent by leveraging the graph clustering structure. 
    At each step, it samples a block of nodes that associate with a dense subgraph identified by a graph clustering algorithm and restricts the neighborhood search within this subgraph.
    
    \noindent\textbf{SGC}~\cite{wu2019sgc} simplifies GCN by removing non-linearities and collapsing weight matrices between consecutive layers.
    Theoretical analysis shows that the simplified model corresponds to a fixed low-pass filter followed by a linear classifier.

    \noindent\textbf{SIGN}~\cite{frasca2020sign} introduces a novel, efficient, and scalable graph deep learning architecture that eliminates the need for graph sampling.
    This method sidesteps the need for graph sampling by using graph convolutional filters of different size.

    \noindent\textbf{GBP}~\cite{chen2020gbp} introduces a scalable GNN that employs a localized bidirectional propagation process involving both feature vectors and the nodes involved in training and testing.
    
    \noindent\textbf{S$^2$GC}~\cite{zhu2021ssgc} introduces a modified Markov Diffusion Kernel for GCN, which strikes a balance between low- and high-pass filters to capture the global and local contexts of each node.

    \noindent\textbf{AGP}~\cite{wang2021agp} proposes a unified randomized algorithm capable of computing various proximity queries and facilitating propagation.
    This method provides a theoretical bounded error guarantee and runs in almost optimal time complexity. 

    \noindent\textbf{GAMLP}~\cite{gamlp} is designed to capture the inherent correlations between different scales of graph knowledge to break the limitations of the enormous size and high sparsity level of graphs hinder their applications under industrial scenarios.

    \noindent\textbf{SEAL}~\cite{zhang2018seal} is learning heuristics from the network. 
    By extracting local subgraphs around target links, SEAL aims to map subgraph patterns to link existence, thus learning suitable heuristics. 
    Based on the heuristic theory, SEAL learns heuristics using a GNN.

    \noindent\textbf{NeoGNN}~\cite{yun2021neognn} learns structural features from the adjacency matrix and estimates overlapped neighborhoods for link prediction. 
    This method generalizes neighborhood overlap-based heuristics and manages multi-hop neighborhoods, enhancing link prediction.

    \noindent\textbf{BUDDY}~\cite{chamberlain2022buddy} propose efficient link prediction with hashing, a novel full-graph GNN that uses subgraph sketches to approximate key components of subgraph GNNs without explicit subgraph construction. 
    For scalability, BUDDY uses feature pre-computation to maintain predictive performance without GPU memory constraints.

    \noindent\textbf{NCNC}~\cite{wang2023ncnc} introduces MPNN-then-SF, an architecture using structural features to guide MPNN’s representation pooling, implemented as a neural common neighbor.
    To mitigate graph incompleteness, NCNC uses a link prediction model to complete the common neighbor structure, achieving neural common neighbor with completion.

    \noindent\textbf{MPLP}~\cite{dong2023mplp} harnesses quasi-orthogonal vectors to estimate link-level structural features while preserving node-level complexities. 
    This technology breaks the limitation of node-level representation and struggles with joint structural features essential for link prediction, like a common neighbor.

    \noindent\textbf{GraphEditor}~\cite{cong2022grapheditor} is an efficient approach for unlearning that supports graph deletion for linear GNN.
    It doesn't require retraining from scratch or access to all training data and ensures exact unlearning, guaranteeing the removal of all corresponding information.

    \noindent\textbf{GUIDE}~\cite{wang2023guide} improves GraphEraser by the graph partitioning with fairness and balance, efficient subgraph repair, and similarity-based aggregation.
    Notably, GUIDE can be efficiently implemented on the inductive graph learning tasks for its low graph partition cost, no matter on computation or structure information.
    
    \noindent\textbf{GraphRevoker}~\cite{zhang2024graphrevoker} is a novel GU framework that maintains model utility better than traditional retraining-based methods. 
    Unlike conventional approaches that partition the training graph into subgraphs, leading to information loss, it employs graph property-aware sharding to preserve graph properties and uses graph contrastive sub-model aggregation for effective prediction. 

    \noindent\textbf{CGU}~\cite{chien2022cgu} introduces the first approach for approximate GU with provable guarantees.
    Their method addresses diverse unlearning requests and evaluates feature mixing during propagation. 
    Notably, CGU only focus on SGC and generalized PageRank extensions.

    \noindent\textbf{CEU}~\cite{wu2023ceu_link} focuses on edge unlearning in GNNs, training a new GNN as if certain edges never existed. 
    CEU updates the pre-trained GNN model parameters in a single step, efficiently removing the influence of specific edges. 
    The authors provide rigorous theoretical guarantees under convex loss function assumptions.

    \noindent\textbf{GIF}~\cite{wu2023gif} incorporates an additional loss term by graph-based influence function, considering structural dependencies, and provides a closed-form solution by efficient estimation of the inverse Hessian matrix for better understanding the unlearning mechanism.

    \noindent\textbf{D2DGN}~\cite{sinha2023d2dgn} is a novel knowledge distillation approach for learning-based GU.
    This framework divides graph knowledge for retention and deletion, using response-based soft targets and feature-based node embeddings while minimizing KL divergence. 
    
    \noindent\textbf{GNNDelete}~\cite{cheng2023gnndelete} is a novel model-agnostic layer-wise operator designed to optimize topology influence in the GU requests.
    This method supervises the unlearning operator by designing a regularization term based on the interactions between graph entities.
    
    \noindent\textbf{UtU}~\cite{tan2024utu_link} targets edge unlearning and claims that current methods effectively remove specific edges but suffer from over-forgetting, leading to performance decline. 
    Based on this, UtU simplifies the process by unlinking the forgotten edges from the graph structure, improving efficiency and preserving performance.

    \noindent\textbf{MEGU}~\cite{li2024megu} is a novel paradigm that simultaneously evolves predictive and unlearning capacities. 
    Specifically, it ensures complementary optimization within a unified training framework, balancing performance and generalization to meet forgetting and prediction.

    \noindent\textbf{ScaleGUN}~\cite{yi2024ScaleGUN} is a scalable and certifiable GU mechanism by lazy local propagation, ensuring certified unlearning in three scenarios.

\begin{table*}[]
\setlength{\abovecaptionskip}{0.2cm}
\setlength{\belowcaptionskip}{-0.2cm}
\caption{Edge-level predictive performance under feature and node unlearning. 
}
\label{tab: sgu edge-level feature and node unlearning}
\begin{tabular}{cc|cccccc}
\midrule[0.3pt]
\multirow{2}{*}{GU ($\downarrow$)} & \multirow{2}{*}{Backbone ($\downarrow$)} & \multicolumn{2}{c}{collab} & \multicolumn{2}{c}{ppa} & \multicolumn{2}{c}{citation2} \\
                                   &                                          & Feature      & Node        & Feature    & Node       & Feature       & Node          \\ \midrule[0.3pt]
\multirow{5}{*}{ScaleGUN}               & SEAL                                & 63.46±0.32   & 64.12±0.44  & 49.86±0.31 & 50.28±0.28 & 85.92±0.40    & 86.71±0.39    \\
                                   & NeoGNN                                   & 58.15±0.46   & 59.33±0.53  & 49.50±0.44 & 49.87±0.45 & 86.38±0.51    & 86.52±0.48    \\
                                   & BUDDY                                    & 66.34±0.65   & 66.85±0.70  & 50.34±0.27 & 50.76±0.36 & 87.87±0.36    & 87.23±0.44    \\
                                   & NCNC                                     & 66.87±0.72   & 67.29±0.65  & 58.79±0.36 & 59.18±0.47 & 88.04±0.39    & 88.40±0.37    \\
                                   & MPLP                                     & 67.32±0.57   & 67.51±0.53  & 61.55±0.52 & 61.72±0.41 & 89.83±0.32    & 90.16±0.30   \\ \midrule[0.3pt]
\multirow{5}{*}{UtU}               & SEAL                                     & 64.92±0.75   & 64.83±0.84  & 50.23±0.62 & 50.75±0.57 & 86.63±0.62    & 87.06±0.57    \\
                                   & NeoGNN                                   & 58.43±0.37   & 59.10±0.52  & 50.42±0.58 & 50.20±0.65 & 86.94±0.84    & 86.83±0.73    \\
                                   & BUDDY                                    & 66.90±0.68   & 67.41±0.61  & 50.56±0.51 & 51.13±0.70 & 88.05±0.50    & 87.71±0.64    \\
                                   & NCNC                                     & 67.28±0.57   & 66.82±0.49  & 59.82±0.75 & 60.26±0.63 & 88.76±0.62    & 89.24±0.58    \\
                                   & MPLP                                     & 67.65±0.83   & 67.56±0.75  & 62.17±0.83 & 62.34±0.72 & 90.21±0.45    & 90.45±0.52    \\ \midrule[0.3pt]
\multirow{5}{*}{SGU}               & SEAL                                     & 66.84±0.63   & 66.52±0.72  & 51.85±0.56 & 52.17±0.62 & 88.72±0.48    & 88.93±0.62    \\
                                   & NeoGNN                                   & 60.11±0.54   & 60.94±0.60  & 52.49±0.60 & 52.32±0.51 & 88.31±0.67    & 88.64±0.58    \\
                                   & BUDDY                                    & 67.76±0.89   & 68.22±0.81  & 52.13±0.47 & 52.68±0.39 & 89.02±0.73    & 89.50±0.69    \\
                                   & NCNC                                     & 68.65±0.52   & 68.54±0.64  & 62.50±0.51 & 62.39±0.50 & 89.85±0.64    & 90.17±0.54    \\
                                   & MPLP                                     & 68.97±0.73   & 69.27±0.78  & 63.58±0.82 & 63.70±0.77 & 90.79±0.55    & 90.86±0.77    \\ \midrule[0.3pt]
\end{tabular}
\end{table*}

\begin{table}[]
\setlength{\abovecaptionskip}{0.2cm}
\setlength{\belowcaptionskip}{-0.2cm}
\caption{Edge-level performance within StealLink.
}
\label{tab: link-specific unlearning}
\begin{tabular}{cc|ccc}
\midrule[0.3pt]
GU ($\downarrow$) & Backbone ($\downarrow$) & collab & ppa   & citation2 \\ \midrule[0.3pt]
\multirow{4}{*}{SEAL}              & GraphRevoker            & 0.508  & OOT & OOM     \\
                  & CEU                     & 0.534  & OOT & OOM     \\
                  & UtU                     & 0.532  & 0.544 & 0.549     \\
                  & SGU                     & 0.519  & 0.532 & 0.528     \\ \midrule[0.3pt]
\multirow{4}{*}{NCNC}              & GraphRevoker            & 0.513  & OOT & OOM     \\
                  & CEU                     & 0.535  & OOT & OOM     \\
                  & UtU                     & 0.538  & 0.553 & 0.547     \\
                  & SGU                     & 0.515  & 0.525 & 0.536     \\ \midrule[0.3pt]
\multirow{4}{*}{MPLP}              & GraphRevoker            & 0.517  & OOT & OOM     \\
                  & CEU                     & 0.531  & OOT & OOM     \\
                  & UtU                     & 0.526  & 0.539 & 0.551     \\
                  & SGU                     & 0.520  & 0.531 & 0.534     \\ \midrule[0.3pt]
\end{tabular}
\end{table}

\subsection{Hyperparameter settings}
\label{appendix: Hyperparameter settings}
    The hyperparameters in the backbones and GU are set according to the original paper if available.
    Otherwise, we perform a hyperparameter search via the Optuna~\cite{akiba2019optuna}.
    Specifically, we explore the optimal shards within the ranges of 20 to 100.
    The weight coefficients of the loss function and other hyperparameters are obtained by means of an interval search from $\{0, 1\}$ or the interval suggested in the original paper. 
    For our proposed SGU, the detailed hyperparameter settings are as follows. 
    To begin with, in the NIM process, we explore the optimal number of graph propagation steps within the range of 3 to 10, while adhering to the original GNN backbone propagation steps ($k$ in Eq.~(\ref{eq: weight-free graph propagation})). 
    Subsequently, we set the default unlearning budget $\mathcal{B}$ to be three times of $|\Delta\mathcal{V}|$ for selecting HIE, and the optimal threshold $\theta$ is searched within the range of 0.5 to 1.
    After that, we construct $\lambda$-flexible entity-specific optimization objectives for efficient fine-tuning. 
    The overall loss function is represented as $\mathcal{L} = \lambda\mathcal{L}_f+(1-\lambda)\mathcal{L}_p$, where $\lambda$ is searched within the range of 0 to 1.
    In addition, we choose Adam as the default optimizer. 
    As for the learning rate, weight decay, and dropout, they exhibit significant variations across GNN backbones, GU strategies, and unlearning scenarios.
    Therefore, we perform separate hyperparameter searches to report the best performance. 
    Notably, for learning-based GU approaches, the training epoch is set to 50, benefiting from the fine-tuning.
    For both node- and link-level scenarios, we split all datasets following the guidelines of recent GU approaches~\cite{cheng2023gnndelete,li2024megu,wu2023ceu_link,tan2024utu_link}, which randomly allocate 80\% of nodes for training and 20\% for testing, and 90\% of edges for training and 10\% for testing.
    To alleviate the randomness and ensure a fair comparison, we repeat each experiment 10 times for unbiased performance.
    Notably, we experiment with multiple GNN backbones in separate experimental modules to validate the generalizability of SGU and avoid complex charts, making the results more reader-friendly.

\subsection{Experiment Environment}
\label{appendix: Experiment Environment}
    Our experiments are conducted on the machine with Intel(R) Xeon(R) Platinum 8468V, and NVIDIA H800 PCIe and CUDA 12.2. 
    The operating system is Ubuntu 20.04.6. 
    As for the software, we use Python 3.8 and Pytorch 2.2.1 for implementation.

\subsection{Link-specific Evaluation}
\label{appendix: Link-specific Evaluation}
    For link-specific scenarios, the evaluation methodologies are as follows:
    (1) \textbf{Model Update}: 
    Inspired by prevalent methods~\cite{wu2023ceu_link}, we employ StealLink~\cite{he2021steallink_attack}, an edge MIA, to empirically evaluate the extent to which the model has forgotten the removed edges. 
    A higher AUC (>0.5) indicates lower unlearning performance, whereas an AUC close to 0.5 implies complete forgetting.
    (2) \textbf{Inference Protection}:
    We directly evaluate the reasoning capability of the modified model by reporting the Non-UE predictions (i.e., link prediction) via the metric suggested by the corresponding benchmark datasets.
    Specifically, according to the suggestion of OGB official documents, we utilize HR@50, HR@100, and MRR to evaluate the predictive performance of ogb-collab, ogb-ppa, and ogb-citation2, respectively.

    To extend SGU to link-specific scenarios, we summarize the following key points: 
    (1) We use the transformation function proposed in Eq.~(\ref{eq: ue label-based forgetting1}) to obtain node-level UEs, which are then processed by NIM to obtain HIE for subsequent entity-based optimization objectives; 
    (2) In prediction-level optimization, we need to replace the node-level cross-entropy loss with a link-specific loss function; 
    (3) In embedding-level optimization, since link-specific GNNs still generate node embeddings for edge representation, we can directly apply the loss formulation of SGU as presented in Sec.\ref{sec: Unlearning Entities Perspective}.

    In addition to Table~\ref{tab: sgu edge-level edge unlearning} presented in Sec.~\ref{sec: Performance Comparison}, we include Table~\ref{tab: sgu edge-level feature and node unlearning} to provide a comprehensive evaluation. 
    Our experimental results indicate that SGU achieves the best performance across all datasets and unlearning requests, demonstrating its robust generalization.
    Notably, CEU lacks scalability because it needs to consider all graph elements contributing to the gradients for optimization. 
    Meanwhile, to evaluate the unlearning performance of link-specific GU methods, we report their unlearning performance in Table~\ref{tab: link-specific unlearning}. 
    We observe that, although partition-based GraphRevoker can achieve sufficient forgetting by retraining specific partitions, the high partition cost results in OOT and OOM errors. 
    Compared to other baselines, SGU excels in forgetting capability, thanks to the reliable HIE identified by NIM and the well-designed optimization objectives.

\begin{figure*}[t]
    \centering
        \setlength{\abovecaptionskip}{0.2cm}
    \setlength{\belowcaptionskip}{-0.15cm}
    \includegraphics[width=\textwidth]{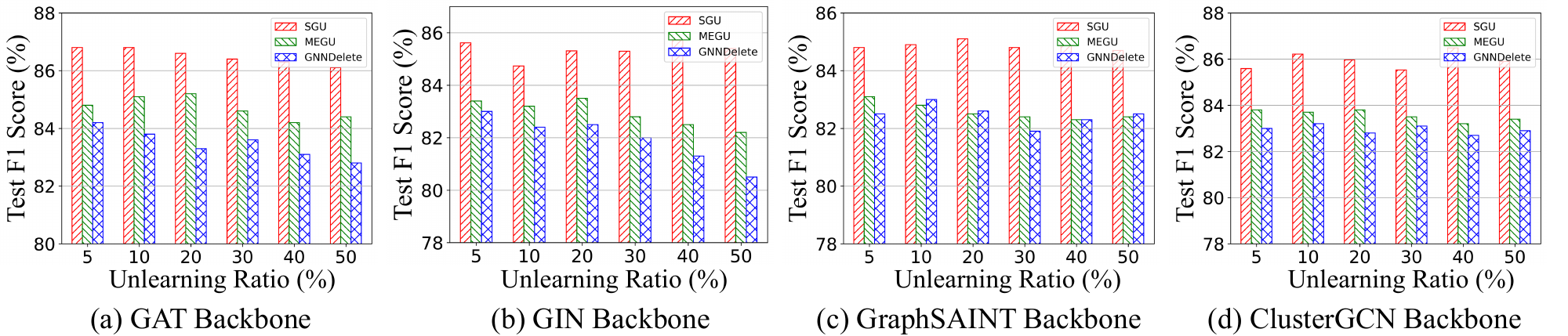}
    \caption{Predictive performance of feature unlearning within different ratios on Cora.}
    \label{fig: ratio_feature_cora}
\end{figure*}

\begin{figure*}[t]
    \centering
        \setlength{\abovecaptionskip}{0.2cm}
    \setlength{\belowcaptionskip}{-0.1cm}
    \includegraphics[width=\textwidth]{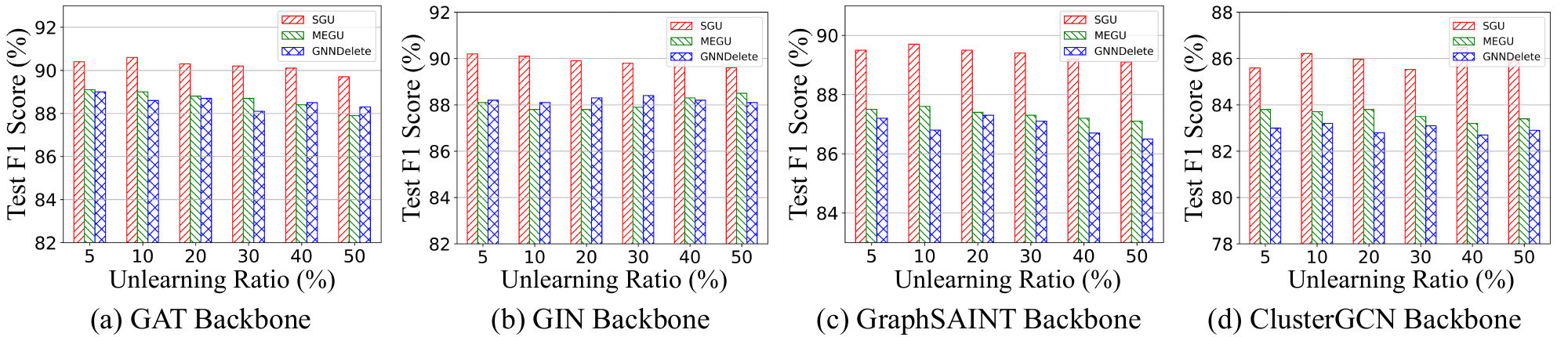}
    \caption{Predictive performance of edge unlearning within different ratios on Photo.}
    \label{fig: ratio_edge_photo}
\end{figure*}

\begin{figure*}[t]
    \centering
        \setlength{\abovecaptionskip}{0.2cm}
    \setlength{\belowcaptionskip}{-0.15cm}
    \includegraphics[width=\textwidth]{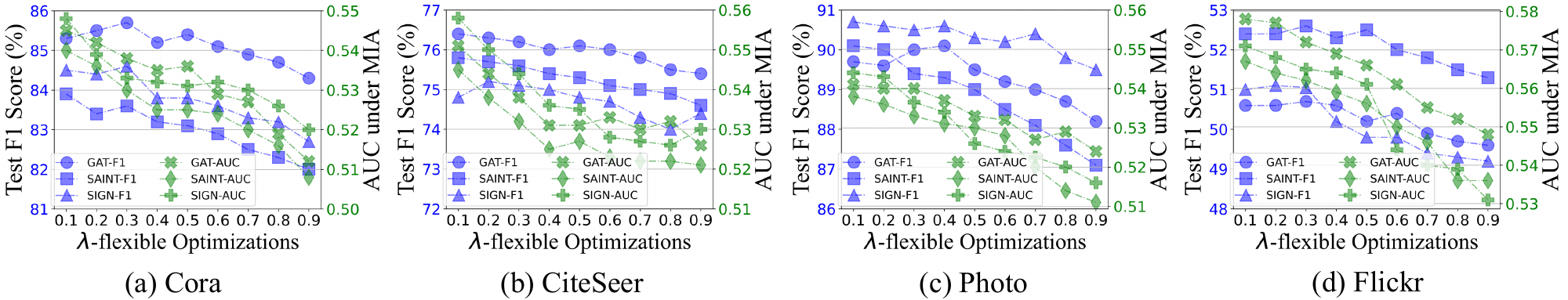}
    \caption{Sensitivity analysis on entity-specify optimization objectives.}
    \label{fig: lambda_optimzation}
\end{figure*}

\subsection{Unlearning Challenges at Different Scales}
\label{appendix: Unlearning Challenges at Different Scales}
    To comprehensively evaluate the efficacy of SGU across varying unlearning scales, in addition to the results provided in Fig.~\ref{fig: ratio_node_ppi} in Sec.~\ref{sec: Robustness Analysis}, we present additional experimental results in Fig.~\ref{fig: ratio_feature_cora} and Fig.~\ref{fig: ratio_edge_photo}. 
    Our experiments reveal that edge unlearning is less impacted than feature and node unlearning on predictive performance.
    In edge unlearning scenarios, we remove the unlearning edges and treat the connected nodes as UE, resulting in more significant unlearning costs. 
    Such discrepancy arises more from the nuanced process associated with node unlearning, where the edge removal disrupts the original topology and causes performance deterioration. 
    In summary, edge unlearning induces a comparatively milder impact on predictive performance for Non-UE compared to other unlearning scenarios.
    Inspired by these findings, we introduce three additional large-scale link-level benchmark datasets and link-specific GNNs as baselines to provide a comprehensive evaluation.


\subsection{$\lambda$-flexible Entity-specific Optimization.}
\label{appendix: flexible Entity-specific Optimization}
    In our proposed NIM and SGU, besides the critical $\mathcal{B}$ and $\theta$ in selecting reliable HIE, the hyperparameter $\lambda$ used to balance forgetting and reasoning loss is also pivotal, as it directly affects unlearning and predictive performance via the fine-tuning. 
    For more details on how $\lambda$ links our proposed optimization objectives, please refer to Appendix~\ref{appendix: Hyperparameter settings}. 
    Therefore, in addition to Fig.~\ref{fig: ratio_node_ppi} provided in Sec.~\ref{sec: Robustness Analysis}, we present an additional sensitivity analysis of SGU to $\lambda$ in Fig.~\ref{fig: lambda_optimzation}.
    
    Notably, in our implementation of linking forgetting and reasoning loss, a larger and smaller $\lambda$ emphasizes SGU's unlearning and predictive performance.
    Based on this, according to the experimental results, we find that in most cases, SGU achieves the best unlearning performance at $\lambda=0.9$, indicated by an AUC closer to 0.5.
    However, $\lambda=0.9$ means that SGU neglects reasoning ability, leading to sub-optimal predictions.
    Therefore, it is important to balance the capabilities of forgetting and reasoning according by adjusting $\lambda$ to actual needs. 

\end{document}